\documentclass{article}
\usepackage{iclr2027_conference,times}

\newcommand{\benchmarkname}{SEER-Repro}

\usepackage{amsmath,amsfonts,bm}

\def\eqref#1{equation~\ref{#1}}

\def\1{\bm{1}}

\DeclareMathAlphabet{\mathsfit}{\encodingdefault}{\sfdefault}{m}{sl}
\SetMathAlphabet{\mathsfit}{bold}{\encodingdefault}{\sfdefault}{bx}{n}

\usepackage{xspace}
\usepackage{booktabs}
\usepackage{multirow}
\usepackage{tabularx}
\usepackage{colortbl}
\usepackage{longtable}
\usepackage{graphicx}
\usepackage{wrapfig}
\usepackage{needspace}
\usepackage{placeins}
\usepackage{tikz}
\usetikzlibrary{arrows.meta,positioning,calc}
\usepackage{etoc}
\usepackage{hyperref}
\usepackage{url}
\usepackage{listings}
\usepackage{algorithm}
\usepackage{algpseudocode}
\usepackage[font=normalsize,justification=raggedright,singlelinecheck=false]{caption}
\lstdefinestyle{seerprompt}{
  basicstyle=\scriptsize\ttfamily,
  columns=fullflexible, keepspaces=true,
  breaklines=true, breakatwhitespace=false,
  breakautoindent=false, breakindent=0pt,
  showstringspaces=false, showspaces=false,
  frame=single, rulecolor=\color{black!35},
  framesep=5pt, xleftmargin=6pt, xrightmargin=6pt,
  aboveskip=8pt, belowskip=12pt, captionpos=t,
  literate={–}{{--}}1 {—}{{---}}1 {×}{{$\times$}}1
           {’}{{'}}1 {‘}{{'}}1 {“}{{``}}1 {”}{{''}}1
}

\newcommand{\reproduced}{\textsc{Reproduced}\xspace}
\newcommand{\close}{\textsc{Close}\xspace}
\newcommand{\notreproduced}{\textsc{NotReproduced}\xspace}

\title{AI-Driven Reproducibility and Analytical Robustness in Scientific Discovery}

\author{%
\textbf{Shicheng Liu\textsuperscript{1}\footnotemark[2], Krista Opsahl-Ong\textsuperscript{2}, Andrew Drozdov\textsuperscript{2}, Julian Killingback\textsuperscript{4,2}} \\
\textbf{Arnav Singhvi\textsuperscript{2}, Ronak Pradeep\textsuperscript{2}, Michael Bendersky\textsuperscript{2}} \\
\textbf{James C. Dickerson\textsuperscript{3}\footnotemark[1], Monica S. Lam\textsuperscript{1}\footnotemark[1]} \\[0.5em]
\normalfont\textsuperscript{1}Department of Computer Science, Stanford University \\
\normalfont\textsuperscript{2}Databricks AI Research \\
\normalfont\textsuperscript{3}Division of Oncology, Stanford University School of Medicine \\
\normalfont\textsuperscript{4}University of Massachusetts Amherst
}

\iclrfinalcopy
\AddToHook{cmd/maketitle/after}{%
  \lhead{Preprint}%
  \begingroup
  \footnotetext[2]{Work done at Databricks AI Research.}%
  \footnotetext[1]{Joint senior authors.}%
  \footnotetext[3]{Correspondence to \texttt{shicheng@cs.stanford.edu}, \texttt{jcdicker@stanford.edu}, and \texttt{lam@cs.stanford.edu}.}%
  \endgroup
}

\begin{document}
\etocdepthtag.toc{main}
\maketitle

\begin{abstract}
Reproducible and analytically robust research findings are essential to scientific progress, but evaluating these properties typically requires costly manual effort. We investigate whether LLM agents can help automate reproduction and robustness analysis, using a medical case study grounded in the Surveillance, Epidemiology, and End Results (SEER) database, a widely used U.S. cancer registry underlying tens of thousands of published studies. First, we introduce \textbf{SEER-Repro}, a benchmark of 20 published SEER-based papers together with the data and resources required to reproduce their findings. Second, we create an AI agent that successfully reproduces 77.4\% of 2,190 atomic claims, while providing executable baselines for further analysis.
Third, we develop an agentic analysis pipeline to test the robustness of the medical findings by identifying analytical choices underlying each finding, proposing defensible alternatives, and rerunning the analysis to determine whether those choices affect the conclusion. Across published papers in \benchmarkname{}, our pipeline identifies potentially consequential analytical choices in 9.5\% of fully reproduced findings and flags them for expert review. These results show how LLM agents can support scientific review by reproducing findings, systematically testing their robustness, and directing expert attention toward potentially consequential analytical choices.

\end{abstract}

\section{Introduction}
\begin{figure}[ht!]
\centering
\definecolor{overviewInk}{HTML}{293C49}
\definecolor{overviewBlue}{HTML}{477E9A}
\definecolor{overviewTeal}{HTML}{398578}
\definecolor{overviewGold}{HTML}{A86E32}
\resizebox{\linewidth}{!}{%
\begin{tikzpicture}[
    font=\rmfamily\fontsize{9}{10.5}\selectfont,
    text=overviewInk,
    flow/.style={-{Latex[length=1.8mm]},draw=overviewInk!65,line width=0.85pt,rounded corners=4pt},
    outline/.style={draw=overviewInk!55,line width=0.7pt,line cap=round,line join=round},
    textline/.style={draw=overviewInk!30,line width=1pt,line cap=round},
    label/.style={align=center,inner sep=0pt},
    heading/.style={label,font=\rmfamily\bfseries\fontsize{10}{11.5}\selectfont},
    legend/.style={label,font=\rmfamily\fontsize{8}{9}\selectfont},
    note/.style={label,font=\rmfamily\fontsize{8.5}{10}\selectfont}
]
\path[use as bounding box] (-0.70,-0.03) rectangle (13.7,5.70);
\node[heading,font=\rmfamily\bfseries\fontsize{12}{14}\selectfont] at (8.13,5.28) {AI-assisted robustness assessment};

\begin{scope}[yshift=-0.25cm]
\draw[draw=overviewTeal!65,dashed,line width=0.7pt,rounded corners=5pt] (-0.62,1.30) rectangle (2.05,5.65);
\begin{scope}[xshift=-0.25cm]
\node[heading,text=overviewTeal,font=\rmfamily\bfseries\fontsize{9}{10.5}\selectfont] at (0.96,5.15) {\benchmarkname{}\\environment};
\begin{scope}[yshift=0.25cm]
\node[heading,font=\rmfamily\bfseries\fontsize{9}{10.5}\selectfont] at (0.96,4.30) {Published paper};
\path[outline,fill=white] (0.42,3.06) -- (0.42,4.03) -- (1.25,4.03) -- (1.54,3.74) -- (1.54,3.06) -- cycle;
\draw[outline] (1.25,4.03) -- (1.25,3.74) -- (1.54,3.74);
\draw[overviewBlue,line width=1.5pt,line cap=round] (0.60,3.80) -- (0.99,3.80);
\foreach \y in {3.56,3.32} \draw[textline] (0.60,\y) -- (1.33,\y);
\end{scope}
\begin{scope}[yshift=0.55cm]
\node[heading,font=\rmfamily\bfseries\fontsize{9}{10.5}\selectfont] at (0.96,2.45) {Data + manuals};
\path[outline,fill=overviewBlue!8] (0.30,1.14) -- (0.30,2.02) arc (180:360:0.50 and 0.15) -- (1.30,1.14) arc (0:-180:0.50 and 0.15);
\draw[outline,fill=white] (0.80,2.02) ellipse (0.50 and 0.15);
\draw[outline] (0.30,1.58) arc (180:360:0.50 and 0.15);
\draw[outline,fill=white,rounded corners=1pt] (1.07,1.04) rectangle (1.65,1.74);
\draw[overviewTeal!65,line width=1pt] (1.16,1.04) -- (1.16,1.74);
\draw[textline] (1.27,1.55) -- (1.49,1.55);
\draw[textline] (1.27,1.36) -- (1.49,1.36);
\end{scope}
\end{scope}
\end{scope}
\draw[flow] (2.17,3.55) -- (2.37,3.55) -- (2.37,2.80) -- (2.63,2.80);
\draw[draw=overviewInk!65,line width=0.85pt,rounded corners=4pt] (2.17,1.90) -- (2.37,1.90) -- (2.37,2.80);

\foreach \left/\mid/\tone/\step in {2.68/4.33/overviewBlue/1,6.48/8.13/overviewTeal/2,10.28/11.93/overviewGold/3} {
    \draw[draw=\tone!30,fill=\tone!6,rounded corners=5pt] (\left,1.62) rectangle ({\left+3.30},4.76);
    \node[circle,fill=\tone,text=white,inner sep=0pt,minimum size=0.38cm,font=\rmfamily\bfseries\fontsize{9}{10}\selectfont] at (\mid,4.76) {\step};
}
\node[heading] at (4.33,4.28) {Extract claims};
\node[heading] at (8.13,4.28) {Reproduce findings};
\node[heading] at (11.93,4.28) {Test robustness};
\node[label] at (4.33,3.68) {Identify findings and\\extract atomic claims};
\node[label] at (8.13,3.68) {Re-run analyses and\\check published claims};
\node[label] at (11.93,3.68) {Test defensible choices;\\compare conclusions};
\draw[flow] (6.00,2.80) -- (6.44,2.80);
\draw[flow] (9.80,2.80) -- (10.24,2.80);

\draw[outline,fill=white,rounded corners=2pt] (3.08,1.93) rectangle (3.84,3.03);
\draw[overviewBlue,line width=1.5pt,line cap=round] (3.25,2.80) -- (3.59,2.80);
\foreach \y in {2.54,2.28} \draw[textline] (3.25,\y) -- (3.65,\y);
\draw[draw=overviewBlue!65,line width=0.7pt] (3.84,2.48) -- (4.12,2.48) (4.12,2.05) -- (4.12,2.91);
\foreach \y/\claim in {2.91/1,2.48/2,2.05/3} {
    \draw[draw=overviewBlue!65,line width=0.7pt] (4.12,\y) -- (4.43,\y);
    \node[draw=overviewBlue!45,fill=white,rounded corners=2pt,minimum width=1.14cm,minimum height=0.31cm,inner sep=1pt,font=\rmfamily\fontsize{8}{9}\selectfont] at (5.00,\y) {Claim \claim};
}

\draw[outline] (6.99,3.04) -- (6.99,2.07) -- (9.31,2.07);
\foreach \x/\h in {7.17/0.40,7.89/0.63,8.61/0.83} {
    \fill[overviewInk!30] (\x,2.08) rectangle ({\x+0.18},{2.08+\h});
    \fill[overviewTeal!80] ({\x+0.22},2.08) rectangle ({\x+0.40},{2.08+\h-0.025});
}
\node[legend] at (8.13,1.85) {\textcolor{overviewInk!30}{\rule{4pt}{4pt}}\,Reported\quad\textcolor{overviewTeal!80}{\rule{4pt}{4pt}}\,Reproduced};

\draw[outline] (10.78,3.04) -- (10.78,2.07) -- (13.10,2.07);
\draw[overviewBlue,line width=1.2pt,line cap=round] (10.90,2.22) .. controls (11.51,2.27) and (12.06,2.89) .. (12.98,2.93);
\draw[overviewGold,dashed,line width=1.2pt,line cap=round] (10.90,2.24) .. controls (11.51,2.36) and (12.06,2.66) .. (12.98,2.63);
\node[legend] at (11.93,1.85) {\textcolor{overviewBlue}{\rule[2pt]{7pt}{0.9pt}}\,Baseline\quad\textcolor{overviewGold}{\rule[2pt]{2.5pt}{0.9pt}\kern2pt\rule[2pt]{2.5pt}{0.9pt}}\,Alternative};

\draw[flow] (11.93,1.60) -- (11.93,1.22);
\begin{scope}[yshift=0.45cm]
\fill[overviewInk!4,rounded corners=4pt] (2.68,-0.46) rectangle (13.58,0.75);
\node[label,anchor=west,font=\rmfamily\bfseries\fontsize{9}{10.5}\selectfont] at (2.96,0.40) {Evidence for expert review};

\foreach \start/\tone/\first in {7.00/overviewBlue/1,9.10/overviewGold/4,11.20/overviewGold/7} {
    \foreach \offset in {0,1,2} {
        \pgfmathtruncatemacro{\score}{\first+\offset}
        \pgfmathsetmacro{\shade}{ifthenelse(\first==7,28,11)}
        \fill[\tone!\shade,rounded corners=1pt] ({\start+0.64*\offset},0.215) rectangle ({\start+0.64*\offset+0.60},0.585);
        \node[legend] at ({\start+0.64*\offset+0.30},0.40) {\score};
    }
}
\node[legend] at (7.94,-0.115) {Numerical\\sensitivity};
\node[legend] at (10.04,-0.115) {Secondary change\\or qualification};
\node[legend] at (12.14,-0.115) {Major effects on\\main conclusions};

\end{scope}
\end{tikzpicture}%
}
\caption{AI-assisted robustness assessment pipeline to support expert review.}
\label{fig:robustness-overview}
\end{figure}
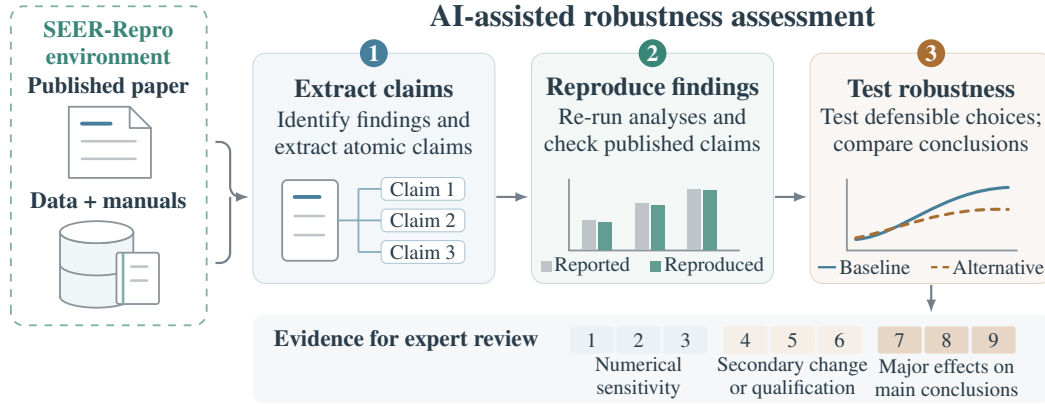

Reproducibility and analytical robustness are the bedrock on which scientific discovery rests.
Reproducibility ensures that research findings can be independently verified, while analytical robustness asks whether conclusions remain stable under reasonable alternative choices. Traditionally, meta-analyses and systematic reviews have addressed these challenges through laborious, manual efforts to reproduce experiments or reanalyze data using alternate specifications \citep{delios2022generalizability,brodeur2026reproducibility}. However, the immense cost and expertise required for such metascientific investigations have limited their scalability.
Scientific claims also depend on the analytical decisions made by their authors \citep{simonsohn2020specification,bertran2026manyai}.
Recent ``many-analyst'' studies in psychology and the social sciences show that independent teams addressing the same hypothesis and dataset can reach conflicting conclusions \citep{silberzahn2018manyanalysts,breznau2022observing,botviniknezer2020variability}. Large language models (LLMs) offer a way to explore this analytical diversity at scale. Importantly, these AI-driven results exhibit notable steerability: shifting the design of the analyst persona or changing the underlying LLM meaningfully alters the distribution of analytic outcomes, even when methodological rigor is maintained throughout \citep{bertran2026manyai,miao2026agenticgarden}.

Does this sensitivity to analytical choices documented in psychological and social sciences extend to medicine, where findings profoundly inform clinical decisions and patient care? Addressing this question is critical, as the validity of medical research underpins clinical decisions that affect countless lives. In this paper, we investigate the potential of AI-driven analysts to enhance both \textit{reproducibility} and \textit{analytical robustness} in scientific discovery, with a particular focus on medical epidemiology using the Surveillance, Epidemiology, and End Results (SEER) database.
Maintained by the National Cancer Institute since 1973, the program integrates population-based cancer-registry records on patient demographics, tumor characteristics, treatment, and survival. In its 2024 usage metrics, NCI reports that 17 core registries cover roughly 48\% of the U.S. population and collect approximately 950,000 incident cases per year \citep{seerprogram,seermetrics2024}. From 1975 through 2024, approximately 25,000 publications used SEER data in their primary analysis, while more than 126,000 publications referenced SEER data~\citep{seermetrics2024}. This scale provides a large body of realistic, high-impact questions, while differences in data and analytical definitions, such as registry coverage and cohort selection, make faithful reconstruction challenging.

We argue that, before LLMs can be trusted to produce new scientific results, they must first demonstrate the ability to accurately reproduce existing scientific studies and assess their analytical robustness. Motivated by this prerequisite, this paper addresses two key research questions:

\textit{First, can we create an AI agent to reproduce published medical epidemiological studies? And if so, what is the reproducibility rate of such papers?} We construct \textbf{\benchmarkname{}}, the first benchmark consisting of 20 original recent or highly cited papers published on the SEER data. We develop an AI agent to reproduce the results of each paper by reading the paper and database documentation, writing and executing the code to access and process the SEER data accordingly. Across 2,190 atomic claims, our pipeline reproduced 77.4\%, while 3.5\% were close and 19.2\% were not reproduced. Error analysis identifies unavailable data as the main barrier to reproduction, despite extensive efforts to recover all required data.

\textit{Second, can we create an AI agent to evaluate the analytical robustness of findings in \benchmarkname{}? And if so, how robust are those findings to reasonable alternative analytical choices?} We formulate the task of \textbf{analytical robustness evaluation} as follows:  Given a research report and access to the underlying data, the task is to investigate if there are well-justified analytical choices that can change the conclusions of the report. We refer to such choices as \textbf{consequential}. Evaluating analytical robustness is challenging because gold-standard judgments do not exist. We therefore investigate three complementary approaches: (1) asking a coding agent to independently analyze the data and comparing its conclusions with those of the published paper; (2) systematically perturbing individual analytical choices through a structured agentic pipeline; and (3) asking coding agents to assess analytical robustness directly. To rate how strongly alternative analyses could change published conclusions, we adapted GRADE~\citep{schunemann2008grade}, an established framework for evaluating clinical evidence that rates the importance of outcomes on a nine-point scale, with scores of 7-9 indicating outcomes critical to decision-making.

Our results show that major effects (scores 7--9) are flagged in 9.5\% of fully reproduced findings (55 of 579). Expert review of 10 flagged cases found that 9 did not establish a consequential change under a defensible alternative, while 1 required further review. Thus, in contrast to the sensitivity reported for the social science datasets studied by \citet{bertran2026manyai} and \citet{miao2026agenticgarden}, our findings suggest that the medical papers in \benchmarkname{} exhibit a high degree of robustness to the analytical alternatives tested—an encouraging sign for the reliability of published medical research.

Overall, our results show how LLM agents can strengthen the scientific process by reproducing findings, systematically testing their robustness, and directing expert attention toward potentially consequential analytical choices.

\section{Related Work}

\subsection{Reproduction and Robustness to Analytical Choice in Scientific Research}

Multiverse and specification-curve analyses examine results across defensible analytical specifications, while many-analyst studies demonstrate that independent teams can reach different conclusions from the same data and question \citep{steegen2016multiverse,simonsohn2020specification,silberzahn2018manyanalysts}. Related efforts test published findings through reproduction and reanalysis: \citet{delios2022generalizability} extend archival analyses to new observations, and \citet{brodeur2026reproducibility} reproduce analyses and conduct robustness checks for 110 economics and political science articles. These studies motivate assessing both whether a finding can be reproduced and whether it survives alternative analytical choices. In this work, we investigate whether an LLM-assisted pipeline can help with this important scientific problem.

Recent work automates robustness exploration with AI agents. \citet{bertran2026manyai} generate analyses with multiple AI analysts and screen their methodological validity with an AI analysis agent. \citet{miao2026agenticgarden} sample agent-generated specifications and estimate how often defensible paths produce claims at least as extreme as the reported finding. Such analytical variation can reflect subjective judgments about how to operationalize a research question. We investigate how this problem manifests in medical epidemiology, where choices about cancer cohorts, outcome definitions, and statistical analyses can alter findings relevant to clinical practice and public health. 

\subsection{Data-Driven Deep Research Benchmarks and Evaluation}

Deep-research benchmarks such as DeepResearchBench, ReportBench, and DeepScholar-Bench assess report quality, evidence coverage, and citation grounding \citep{yang2025deepresearchbench,li2025reportbench,patel2025deepscholarbench}. Data-oriented benchmarks evaluate insight generation and executable scientific workflows, including InsightBench, DiscoveryBench, and ScienceAgentBench \citep{sahu2024insightbench,majumder2025discoverybench,chen2025scienceagentbench}. BLADE additionally evaluates conceptual variables, transformations, and models across multiple valid analyses \citep{gu2024blade}. DataSTORM, Hunt, and DAgent extend this setting to database investigation and report generation \citep{liu2026datastorm,liu2026hunt,xu2025dagent}. Our evaluation asks whether a report's conclusions survive defensible changes to the analytical choices that produced them, using executed alternatives linked to individual findings. This report-level assessment of analytical robustness is not covered by these prior benchmarks and evaluations.

\section{Methodology}

\subsection{Automated Pipeline for Reproducing Published Findings}
\label{sec:reproduction}

In this section, we provide an overview of our proposed agent for reproducing published scientific findings. Our process starts with the paper $p$ that we are interested in reproducing:

\textbf{Extract findings and atomic claims}: First, we extract the empirical findings $F_p=\{f_1,\ldots,f_m\}$. These findings represent the core contributions of the paper and underpin the conclusion it draws. We exclude sections of the paper pertaining to related work, background context, or methodology. 
Each finding $f_i$ is then decomposed into a set of atomic claims $A_{i}=\{a_{ij}\}_{j=1}^{n_i}$. One atomic claim involves one specific numerical or qualitative target while one finding could group multiple atomic claims together. For instance, ``Among women with adenoid cystic carcinoma of the breast, the mean age at diagnosis was 63 years, the median was 62 years, and the range was 33--97 years'' is a finding, while its individual targets such as ``Among women with adenoid cystic carcinoma of the breast, the mean age at diagnosis was 63 years'' (and similarly for ``median age: 62 years'' and ``age range: 33--97 years'') each become a separate atomic claim.

\textbf{Reproduce findings}: We then ask a coding agent to reproduce each finding $f_i$ using the original paper $p$, the data used by the paper, and any additional documentation for the data. The agent identifies the relevant data, writes analysis code, and executes it to produce evidence $e_i$ for whether the finding can be validated or not. The evidence $e_i$ includes the output of any analytical code and its results.

\textbf{Grade individual claims}: Finally, for each atomic claim and evidence pair $(a_{ij},e_i)$, a claim verification agent checks the method, extracts the reproduced result, and assesses the quantitative or qualitative agreement with the known target values in $a_{ij}$ to produce a final verdict $v_{ij}\in\{\reproduced,\close,\notreproduced\}$. For numerical targets, we consider a claim reproduced if the relative error is at most 5\%, while errors greater than 5\% and at most 10\% are counted as close. Qualitative claims are deemed either reproduced or not reproduced. The full protocol for grading is presented in Appendix~\ref{app:benchmark-construction}.

\subsection{Analytical Robustness Analysis}

The reproduction artifacts, including the exact code derived, then serve as a foundation for analytic robustness evaluation. Our overall approach uses 3 complementary methods illustrated in Figure~\ref{fig:analysis-pipeline}.
\label{sec:analysis}

\subsubsection{Independent Report Generation}
\label{sec:independent-reports}

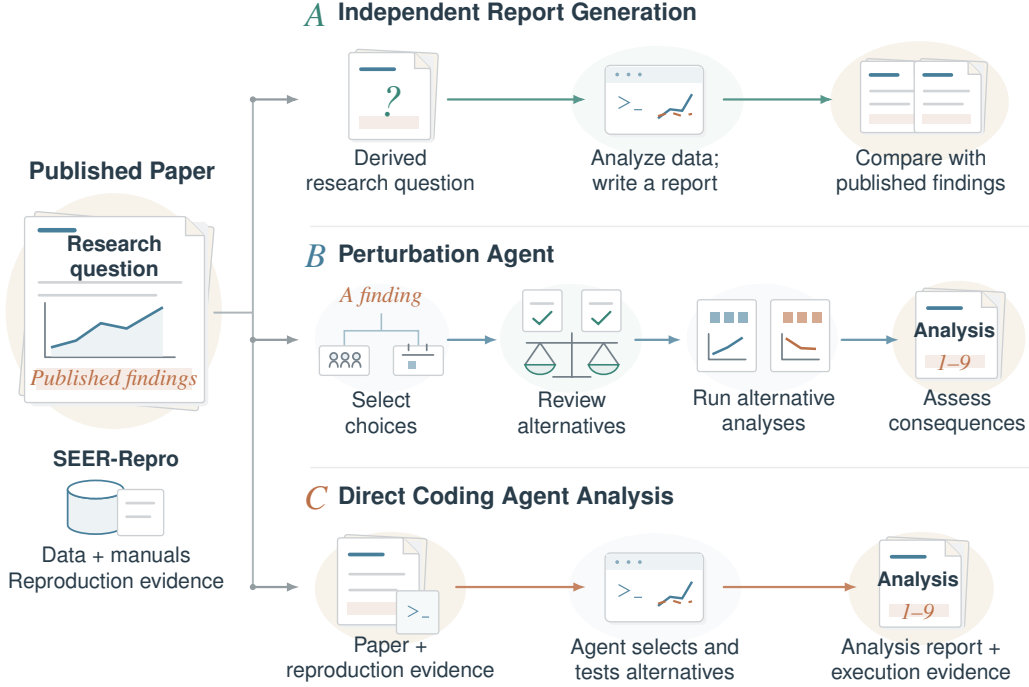
\begin{figure}[!t]
\centering
\definecolor{analysisInk}{HTML}{293C49}
\definecolor{analysisBlue}{HTML}{477E9A}
\definecolor{analysisTeal}{HTML}{398578}
\definecolor{analysisOrange}{HTML}{BB7047}
\definecolor{analysisMuted}{HTML}{77838A}
\definecolor{analysisCream}{HTML}{F7F3EB}
\begin{tikzpicture}[
    font=\sffamily\fontsize{8}{9.5}\selectfont,
    ink/.style={draw=analysisInk!65,line width=0.65pt,line cap=round,line join=round},
    paper/.style={draw=analysisInk!24,fill=white,rounded corners=1pt,line width=0.5pt},
    flow/.style={-{Latex[length=1.7mm]},draw=analysisBlue!80,line width=0.85pt,line cap=round},
    labeltext/.style={inner sep=0pt,align=center,text=analysisInk},
    small/.style={labeltext,font=\sffamily\fontsize{8}{9.5}\selectfont},
    title/.style={labeltext,anchor=west,font=\sffamily\bfseries\fontsize{9}{11}\selectfont},
    rule/.style={draw=analysisInk!22,line width=1pt,line cap=round},
    pathletter/.style={labeltext,font=\rmfamily\itshape\fontsize{13}{15}\selectfont},
    note/.style={labeltext,font=\rmfamily\itshape\fontsize{8.5}{10}\selectfont},
    pics/report/.style={code={
        \begin{scope}[rotate=-3]
        \draw[paper,fill=analysisCream] (-0.49,-0.49) rectangle (0.60,0.66);
        \end{scope}
        \fill[analysisInk!7,rounded corners=1pt] (-0.51,-0.55) rectangle (0.57,0.60);
        \draw[paper] (-0.55,-0.51) -- (-0.55,0.64) -- (0.32,0.64) -- (0.53,0.43) -- (0.53,-0.51) -- cycle;
        \draw[draw=analysisInk!24,fill=analysisCream,line width=0.5pt] (0.32,0.64) -- (0.32,0.43) -- (0.53,0.43);
        \draw[analysisBlue,line width=1.4pt,line cap=round] (-0.39,0.42) -- (0.05,0.42);
        \draw[rule] (-0.39,0.16) -- (0.33,0.16);
        \draw[rule] (-0.39,-0.05) -- (0.23,-0.05);
        \fill[analysisOrange!13] (-0.42,-0.37) rectangle (0.37,-0.21);
    }},
    pics/agent/.style={code={
        \fill[analysisInk!7,rounded corners=2pt] (-0.62,-0.48) rectangle (0.68,0.43);
        \draw[paper,rounded corners=2pt] (-0.66,-0.44) rectangle (0.64,0.47);
        \fill[analysisBlue!10] (-0.64,0.25) rectangle (0.62,0.45);
        \foreach \x in {-0.50,-0.35,-0.20} \fill[analysisBlue!65] (\x,0.35) circle (0.025);
        \node[labeltext,text=analysisBlue,font=\ttfamily\fontsize{10}{11}\selectfont] at (-0.34,-0.04) {>\_};
        \draw[analysisBlue,line width=0.9pt] (0.04,-0.23) -- (0.20,-0.11) -- (0.36,-0.13) -- (0.49,0.09);
        \draw[analysisOrange,dashed,line width=0.8pt] (0.04,-0.20) -- (0.20,-0.15) -- (0.36,-0.20) -- (0.49,-0.17);
    }}
]
\path[use as bounding box] (0,0) rectangle (13.6,10.05);

\fill[analysisCream] (1.32,5.70) ellipse (1.43 and 1.56);
\node[title] at (0.19,7.53) {Published Paper};
\begin{scope}
\draw[paper,fill=analysisCream,rotate around={-5:(1.35,5.74)}] (0.17,4.48) rectangle (2.55,6.92);
\fill[analysisInk!7,rounded corners=2pt] (0.21,4.44) rectangle (2.57,6.88);
\draw[paper] (0.14,4.49) -- (0.14,6.95) -- (2.27,6.95) -- (2.52,6.70) -- (2.52,4.49) -- cycle;
\draw[draw=analysisInk!24,fill=analysisCream,line width=0.5pt] (2.27,6.95) -- (2.27,6.70) -- (2.52,6.70);
\draw[analysisBlue,line width=1.7pt,line cap=round] (0.34,6.77) -- (0.80,6.77);
\node[small,font=\sffamily\bfseries\fontsize{8}{9.5}\selectfont] at (1.33,6.40) {Research\\question};
\draw[rule] (0.33,6.08) -- (2.21,6.08);
\draw[rule] (0.33,5.91) -- (1.88,5.91);
\draw[ink] (0.41,5.81) -- (0.41,5.09) -- (2.13,5.09);
\fill[analysisBlue!10] (0.49,5.10) -- (0.49,5.22) -- (0.82,5.31) -- (1.15,5.54) -- (1.49,5.47) -- (1.97,5.75) -- (1.97,5.10) -- cycle;
\draw[analysisBlue,line width=1pt] (0.49,5.22) -- (0.82,5.31) -- (1.15,5.54) -- (1.49,5.47) -- (1.97,5.75);
\fill[analysisOrange!13] (0.29,4.66) rectangle (2.35,4.95);
\node[note,text=analysisOrange] at (1.32,4.80) {Published findings};
\end{scope}

\node[small,font=\sffamily\bfseries\fontsize{8}{10}\selectfont] at (1.35,3.72) {\benchmarkname{}};
\path[draw=analysisBlue,fill=analysisBlue!10,line width=0.65pt] (0.71,2.87) -- (0.71,3.34) arc (180:360:0.37 and 0.12) -- (1.45,2.87) arc (0:-180:0.37 and 0.12);
\draw[draw=analysisBlue,fill=white,line width=0.65pt] (1.08,3.34) ellipse (0.37 and 0.12);
\draw[paper] (1.37,2.73) rectangle (1.98,3.34);
\draw[rule] (1.49,3.16) -- (1.85,3.16);
\draw[rule] (1.49,2.98) -- (1.80,2.98);
\node[small] at (1.35,2.28) {Data + manuals\\Reproduction evidence};

\draw[analysisMuted!60,line width=0.75pt] (2.64,5.68) -- (3.16,5.68);
\draw[analysisMuted!60,line width=0.75pt] (3.16,2.05) -- (3.16,8.50);
\foreach \y in {8.50,5.32,2.05} {
    \fill[analysisMuted!80] (3.16,\y) circle (1.15pt);
    \draw[flow,draw=analysisMuted!80] (3.16,\y) -- (3.78,\y);
}
\draw[analysisInk!10,line width=0.55pt] (3.92,6.82) -- (13.43,6.82);
\draw[analysisInk!10,line width=0.55pt] (3.92,3.63) -- (13.43,3.63);

\node[pathletter,text=analysisTeal] at (3.99,9.63) {A};
\node[title] at (4.28,9.63) {Independent Report Generation};
\fill[analysisTeal!5] (8.51,8.48) ellipse (1.14 and 0.73);
\fill[analysisCream] (11.98,8.47) ellipse (1.18 and 0.80);
\pic at (4.98,8.48) {report};
\fill[white] (4.54,8.34) rectangle (5.35,8.74);
\node[labeltext,text=analysisTeal,font=\rmfamily\itshape\fontsize{17}{19}\selectfont] at (4.98,8.48) {?};
\pic at (8.48,8.48) {agent};
\begin{scope}[shift={(11.65,8.48)},scale=0.82,transform shape]
\pic {report};
\end{scope}
\begin{scope}[shift={(12.36,8.48)},scale=0.82,transform shape]
\pic {report};
\end{scope}
\draw[flow,draw=analysisTeal!85] (5.74,8.50) -- (7.57,8.50);
\draw[flow,draw=analysisTeal!85] (9.39,8.50) -- (10.84,8.50);
\node[small] at (4.98,7.54) {Derived\\research question};
\node[small] at (8.48,7.54) {Analyze data;\\write a report};
\node[small] at (12.00,7.54) {Compare with\\published findings};

\node[pathletter,text=analysisBlue] at (3.99,6.43) {B};
\node[title] at (4.28,6.43) {Perturbation Agent};
\fill[analysisBlue!4] (4.87,5.34) ellipse (0.89 and 0.72);
\fill[analysisTeal!5] (7.37,5.34) ellipse (0.94 and 0.72);
\fill[analysisBlue!4] (9.90,5.34) ellipse (1.00 and 0.72);
\fill[analysisCream] (12.50,5.34) ellipse (0.88 and 0.76);

\node[note,text=analysisOrange] at (4.86,5.89) {A finding};
\draw[analysisBlue!50,line width=0.6pt] (4.86,5.67) -- (4.86,5.44) (4.37,5.19) -- (4.37,5.44) -- (5.35,5.44) -- (5.35,5.19);
\foreach \x in {4.37,5.35} \draw[paper] ({\x-0.33},4.85) rectangle ({\x+0.33},5.24);
\foreach \x in {4.23,4.38,4.53} {
    \draw[ink,fill=analysisBlue!12] (\x,5.10) circle (0.047);
    \draw[ink] ({\x-0.055},4.96) .. controls ({\x-0.055},5.06) and ({\x+0.055},5.06) .. ({\x+0.055},4.96);
}
\draw[ink] (5.14,5.12) -- (5.55,5.12);
\draw[ink] (5.22,5.18) -- (5.22,5.26) (5.46,5.18) -- (5.46,5.26);
\fill[analysisBlue!50] (5.21,4.94) rectangle (5.31,5.04);

\foreach \x in {6.99,7.77} {
    \draw[paper] ({\x-0.26},5.43) rectangle ({\x+0.26},5.98);
    \draw[rule] ({\x-0.15},5.80) -- ({\x+0.14},5.80);
    \draw[analysisTeal,line width=0.85pt,line cap=round] ({\x-0.12},5.60) -- ({\x-0.04},5.53) -- ({\x+0.12},5.69);
}
\draw[ink] (7.38,5.41) -- (7.38,4.83) (7.13,4.83) -- (7.63,4.83) (6.78,5.28) -- (7.98,5.28);
\foreach \x in {6.96,7.80} {
    \draw[ink] (\x,5.28) -- ({\x-0.20},4.99) (\x,5.28) -- ({\x+0.20},4.99);
    \path[draw=analysisTeal,fill=analysisTeal!12,line width=0.6pt] ({\x-0.20},4.99) .. controls ({\x-0.14},4.82) and ({\x+0.14},4.82) .. ({\x+0.20},4.99) -- cycle;
}

\foreach \x/\tone in {9.43/analysisBlue,10.40/analysisOrange} {
    \draw[paper] ({\x-0.39},4.90) rectangle ({\x+0.39},5.85);
    \foreach \dx in {-0.22,-0.03,0.16} \fill[\tone!60] ({\x+\dx},5.55) rectangle ({\x+\dx+0.12},5.69);
    \draw[ink] ({\x-0.24},5.40) -- ({\x-0.24},5.06) -- ({\x+0.25},5.06);
}
\draw[analysisBlue,line width=0.95pt] (9.24,5.13) -- (9.44,5.22) -- (9.64,5.35);
\draw[analysisOrange,line width=0.95pt] (10.21,5.34) -- (10.41,5.21) -- (10.61,5.20);

\pic at (12.48,5.32) {report};
\fill[white] (12.05,5.17) rectangle (12.86,5.60);
\node[small,font=\sffamily\bfseries\fontsize{7.5}{9}\selectfont] at (12.44,5.42) {Analysis};
\node[note,text=analysisOrange] at (12.43,5.02) {1--9};
\draw[flow] (5.74,5.32) -- (6.40,5.32);
\draw[flow] (8.22,5.32) -- (8.83,5.32);
\draw[flow] (10.95,5.32) -- (11.71,5.32);
\node[small] at (4.86,4.36) {Select\\choices};
\node[small] at (7.38,4.36) {Review\\alternatives};
\node[small] at (9.92,4.36) {Run alternative\\analyses};
\node[small] at (12.48,4.36) {Assess\\consequences};

\node[pathletter,text=analysisOrange] at (3.99,3.23) {C};
\node[title] at (4.28,3.23) {Direct Coding Agent Analysis};
\fill[analysisCream] (4.98,2.03) ellipse (1.00 and 0.75);
\fill[analysisBlue!4] (8.48,2.03) ellipse (1.14 and 0.73);
\fill[analysisCream] (12.00,2.03) ellipse (0.96 and 0.75);
\pic at (4.83,2.06) {report};
\draw[paper,fill=analysisBlue!5] (5.06,1.43) rectangle (5.62,2.00);
\node[labeltext,text=analysisBlue,font=\ttfamily\fontsize{9}{10}\selectfont] at (5.34,1.72) {>\_};
\pic at (8.48,2.03) {agent};
\pic at (12.00,2.03) {report};
\fill[white] (11.57,1.88) rectangle (12.38,2.31);
\node[small,font=\sffamily\bfseries\fontsize{7.5}{9}\selectfont] at (11.96,2.13) {Analysis};
\node[note,text=analysisOrange] at (11.95,1.73) {1--9};
\draw[flow,draw=analysisOrange!85] (5.85,2.05) -- (7.57,2.05);
\draw[flow,draw=analysisOrange!85] (9.39,2.05) -- (11.16,2.05);
\node[small] at (4.98,1.10) {Paper +\\reproduction evidence};
\node[small] at (8.48,1.10) {Agent selects and\\tests alternatives};
\node[small] at (12.00,1.10) {Analysis report +\\execution evidence};
\end{tikzpicture}
\caption{AI-driven analytical robustness with 3 complementary methods discussed in Sec~\ref{sec:analysis}}
\label{fig:analysis-pipeline}
\end{figure}

Our first method asks whether an independent analysis of the same research question would reach a conclusion that contradicts the original paper (Figure~\ref{fig:analysis-pipeline} (A)). For each paper $p$, a coding agent receives a research question specifying the topic and scope derived from $p$, together with access to relevant data and additional documentation about the data, if available, and is tasked with independently answering the research question. The agent chooses its analytical approach, writes and executes code, and produces a research report $r$. We then judge whether the conclusions in $r$ contradict those in $p$. Appendix~\ref{app:llm-research-reports} documents the report-generation setup and the prompts for all 20 research questions.

\subsubsection{Perturbation Agent}
\label{sec:fixed-analysis} 

Our second method uses the Perturbation Agent illustrated in Figure~\ref{fig:analysis-pipeline} (B) and Algorithm~\ref{alg:fixed-analysis}. Appendix~\ref{app:analysis-implementation} details the implementation of each component, including prompts.

\begin{algorithm}[t]
\caption{Perturbation Agent}
\label{alg:fixed-analysis}
\small
\algrenewcommand{\algorithmicrequire}{\textbf{Input:}}
\algrenewcommand{\algorithmicensure}{\textbf{Output:}}
\begin{algorithmic}[1]
\Require Paper $p$, reproduced findings $F^{\mathrm{rep}}_p$, reproduction evidence $E_p$, LLM-generated reports $\mathcal{R}_p$ from Sec~\ref{sec:independent-reports}
\Ensure A list of outcomes $y_{k}(s)\in\texttt{AnalysisOutcome}$ and an associated 1--9 score for $y_k(s) = \texttt{DirectChange}$
\State $\mathcal{C}_p\gets\Call{SelectAnalyticalChoices}{p,\mathcal{R}_p,F^{\mathrm{rep}}_p,E_p}$
\State $\mathcal{K}_p\gets\Call{ProposeAlternatives}{p,\mathcal{R}_p,\mathcal{C}_p,E_p}$
\For{each choice--alternative comparison $(k,s)\in\mathcal{K}_p$}
    \State $e_{k}(s) \gets \Call{ExecuteAnalyses}{k,s}$.
    \State $y_{k}(s) \gets \Call{AssessConsequences}{k,s,e_{k}(s)}$.
    \State If $y_{k}(s)=\texttt{DirectChange}$, assign a 1--9 consequentiality score.
\EndFor
\State \Return comparison outcomes and consequentiality scores.
\end{algorithmic}
\end{algorithm}

\textbf{Eligible findings.} For a published paper $p$, we start from the findings reproduced according to Section~\ref{sec:reproduction}:
\begin{equation}
 F^{\mathrm{rep}}_p=\{f_i \in F_p:v_{ij}\text{ is reproduced for every }a_{ij}\in A_{i}\}.
 \label{eq:analysis-eligible}
\end{equation}
Only fully reproduced findings enter the robustness analysis: their reconstructed results have passed the reproduction criteria in Section~\ref{sec:reproduction}, and their reproduction code provides the baseline for testing alternative analytical choices.

\textbf{$\textsc{SelectAnalyticalChoices}$}: First, the pipeline identifies analytical choices made in $p$ that contribute to the known outcome for the findings $F^{\mathrm{rep}}_p$, such as cohort definitions, endpoints, or statistical models. The set $\mathcal{R}_p$ of independent reports generated for $p$ is used as an additional reference.
These choices are then ordered by scientific importance and feasibility to select up to 20 choices which form the final set of choices $\mathcal{C}_p$.

\textbf{$\textsc{ProposeAlternatives}$}: Next, the pipeline establishes how the paper $p$ implements each selected choice in $\mathcal{C}_p$, using the reproduction evidence $E_p=\{e_i\}_{i=1}^{m}$, where $e_i$ contains the evidence for finding $f_i$. It then proposes defensible alternatives. The resulting set $\mathcal{K}_p$ contains comparisons $(k,s)$, where each $k\in\mathcal{C}_p$ is a selected analytical choice and $s$ is a proposed alternative setting.

\textbf{$\textsc{ExecuteAnalyses}$}: For each comparison $(k,s)$, we run the baseline unchanged and validate it against the available evidence. We then apply the alternative choice while holding the remaining analysis fixed.

\textbf{$\textsc{Assess Consequences}$}: Finally, the pipeline assigns each comparison an outcome $y_{k}(s)$ from the enumerated type
\begin{equation}
\begin{aligned}
\texttt{AnalysisOutcome}=\{&\texttt{DirectChange},\ \texttt{Qualification},\\
&\texttt{StableUnderTest},\ \texttt{Inconclusive},\\
&\texttt{Unavailable},\ \texttt{NotTested}\}.
\end{aligned}
\label{eq:analysis-decision}
\end{equation}

\texttt{DirectChange} denotes a change to a stated finding; \texttt{Qualification} narrows its scope or evidentiary strength without directly changing its conclusion; and \texttt{StableUnderTest} means the completed comparison does not meet the criterion for a consequential change. \texttt{Inconclusive} means the evidence cannot support a verdict, \texttt{Unavailable} indicates missing required inputs or baseline evidence, and \texttt{NotTested} records comparisons excluded before execution.

The outcome label only describes what happens to the tested finding. For \texttt{DirectChange} outcomes, we additionally grade the effect on the source's conclusions from 1 to 9, using a scale adapted from GRADE's outcome-importance framework \citep{schunemann2008grade}. Scores 1--3 indicate quantitative sensitivity without substantive interpretive change, 4--6 indicate qualifications or changes to secondary findings, and 7--9 indicate major qualification, loss of support, or reversal of a principal conclusion.

\subsubsection{Direct Coding Agent Analysis}
\label{sec:direct-analysis}
Our third method asks a coding agent to identify and test consequential analytical choices directly (Figure~\ref{fig:analysis-pipeline} (C)). For each paper, the agent receives the paper $p$, the available reproduction evidence $E_p$, and access to the data and data documentation when available. We then instruct the agent to select scientifically defensible alternatives addressing the same research question, validate the baseline analysis, and, where feasible, execute comparisons that vary one choice while holding the remaining analysis fixed (with Prompt~\ref{prompt:one-shot-consequential-analysis}). It produces a report describing the choices tested, the supporting evidence, and their effects on the source's conclusions. After the analysis, we grade all output choices using the same 1--9 scale as in Section~\ref{sec:fixed-analysis}.

\section{Experiments}
\label{sec:results}
\label{sec:analysis-experiments}

\begin{table}[!t]
\centering
\caption{Atomic reproduction outcomes for the 20-paper \benchmarkname{} benchmark using GPT-5.6 Sol. Appendix Table~\ref{tab:seer-twenty-reproduction-counts} gives exact outcome counts and percentages; Appendix~\ref{app:paper-titles} lists full paper titles; Appendix Table~\ref{tab:seer-twenty-reproduction-breakdown} separates quantitative and qualitative claims.}\label{tab:seer-twenty-reproduction}
\scriptsize
\setlength{\baselineskip}{0.94\baselineskip}
\definecolor{reprobar}{HTML}{B8D8D2}
\definecolor{closebar}{HTML}{E8D5A8}
\definecolor{unreprobar}{HTML}{D3D7DC}
\setlength{\tabcolsep}{2.0pt}
\newlength{\outcomespan}
\setlength{\outcomespan}{\dimexpr 0.350\linewidth-10\tabcolsep\relax}
\newcommand{\fixedtablecell}[2]{\parbox[c]{\linewidth}{#1 #2}}
\newcommand{\stackedoutcomes}[6]{%
  \multicolumn{3}{m{\outcomespan}@{}}{%
    \parbox[c][16pt][c]{\linewidth}{%
      \pgfmathsetlengthmacro{\reproducedwidth}{#1/(#1+#3+#5)*\linewidth}%
      \pgfmathsetlengthmacro{\closewidth}{#3/(#1+#3+#5)*\linewidth}%
      \hbox{%
        \rlap{\color{reprobar}\rule{\reproducedwidth}{10pt}}%
        \makebox[\reproducedwidth][c]{\raisebox{2.5pt}{\color{black}#1 (#2\%)}}%
        {\color{closebar}\rule{\closewidth}{10pt}}%
        {\color{unreprobar}\rule{\dimexpr\linewidth-\reproducedwidth-\closewidth\relax}{10pt}}}%
    }%
  }%
}
\renewcommand{\arraystretch}{1.0}
\begin{tabularx}{\linewidth}{@{}>{\raggedright\arraybackslash}m{0.303\linewidth}>{\raggedright\arraybackslash}m{0.185\linewidth}>{\raggedleft\arraybackslash}m{0.045\linewidth}>{\raggedleft\arraybackslash}m{0.062\linewidth}>{\raggedleft\arraybackslash}m{0.055\linewidth}*{3}{>{\raggedleft\arraybackslash}X}@{}}
\toprule
Paper (shorter name) & Journal (year) & Cit. & Findings & \shortstack{Atomic\\claims} & \multicolumn{3}{m{\outcomespan}@{}}{\hbox to \linewidth{\shortstack[r]{\textcolor{reprobar}{\rule{10pt}{3pt}}\\Reproduced}\hfill\shortstack[r]{\textcolor{closebar}{\rule{10pt}{3pt}}\\Close}\hfill\shortstack[r]{\textcolor{unreprobar}{\rule{10pt}{3pt}}\\Not reproduced}}} \\
\midrule
\rowcolor{black!4}
\fixedtablecell{\raggedright}{Bone sarcoma survival by tumor site~\citep{hu2021bone}} & \fixedtablecell{\raggedright}{Cancers (2021)} & \fixedtablecell{\raggedleft}{37} & \fixedtablecell{\raggedleft}{56} & \fixedtablecell{\raggedleft}{188} & \stackedoutcomes{187}{99.5}{0}{0.0}{1}{0.5} \\
\fixedtablecell{\raggedright}{Sex disparities in cancer incidence~\citep{cook2009sex}} & \fixedtablecell{\raggedright}{Cancer Epidemiol. Biomarkers Prev. (2009)} & \fixedtablecell{\raggedleft}{466} & \fixedtablecell{\raggedleft}{36} & \fixedtablecell{\raggedleft}{151} & \stackedoutcomes{140}{92.7}{1}{0.7}{10}{6.6} \\
\rowcolor{black!4}
\fixedtablecell{\raggedright}{Inflammatory breast cancer incidence and survival~\citep{hance2005inflammatory}} & \fixedtablecell{\raggedright}{J. Natl. Cancer Inst. (2005)} & \fixedtablecell{\raggedleft}{702} & \fixedtablecell{\raggedleft}{40} & \fixedtablecell{\raggedleft}{131} & \stackedoutcomes{119}{90.8}{2}{1.5}{10}{7.6} \\
\fixedtablecell{\raggedright}{Thyroid cancer trends~\citep{davies2014thyroid}} & \fixedtablecell{\raggedright}{JAMA Otolaryngol. Head Neck Surg. (2014)} & \fixedtablecell{\raggedleft}{1,628} & \fixedtablecell{\raggedleft}{33} & \fixedtablecell{\raggedleft}{91} & \stackedoutcomes{80}{87.9}{1}{1.1}{10}{11.0} \\
\rowcolor{black!4}
\fixedtablecell{\raggedright}{Age at oropharyngeal cancer diagnosis~\citep{cline2020oropharyngeal}} & \fixedtablecell{\raggedright}{Cancers (2020)} & \fixedtablecell{\raggedleft}{29} & \fixedtablecell{\raggedleft}{75} & \fixedtablecell{\raggedleft}{154} & \stackedoutcomes{129}{83.8}{9}{5.8}{16}{10.4} \\
\fixedtablecell{\raggedright}{Adenoid cystic carcinoma of the breast~\citep{ghabach2010adenoid}} & \fixedtablecell{\raggedright}{Breast Cancer Res. (2010)} & \fixedtablecell{\raggedleft}{188} & \fixedtablecell{\raggedleft}{51} & \fixedtablecell{\raggedleft}{128} & \stackedoutcomes{103}{80.5}{2}{1.6}{23}{18.0} \\
\rowcolor{black!4}
\fixedtablecell{\raggedright}{Adult diffuse glioma incidence~\citep{li2018glioma}} & \fixedtablecell{\raggedright}{Cancer Med. (2018)} & \fixedtablecell{\raggedleft}{72} & \fixedtablecell{\raggedleft}{60} & \fixedtablecell{\raggedleft}{111} & \stackedoutcomes{89}{80.2}{4}{3.6}{18}{16.2} \\
\fixedtablecell{\raggedright}{Childhood cancer incidence and mortality~\citep{linet1999childhood}} & \fixedtablecell{\raggedright}{J. Natl. Cancer Inst. (1999)} & \fixedtablecell{\raggedleft}{438} & \fixedtablecell{\raggedleft}{52} & \fixedtablecell{\raggedleft}{144} & \stackedoutcomes{113}{78.5}{6}{4.2}{25}{17.4} \\
\rowcolor{black!4}
\fixedtablecell{\raggedright}{Head and neck mucosal melanoma incidence~\citep{marcus2012mucosal}} & \fixedtablecell{\raggedright}{J. Skin Cancer (2012)} & \fixedtablecell{\raggedleft}{83} & \fixedtablecell{\raggedleft}{20} & \fixedtablecell{\raggedleft}{54} & \stackedoutcomes{42}{77.8}{1}{1.9}{11}{20.4} \\
\fixedtablecell{\raggedright}{Esophageal adenocarcinoma incidence and mortality~\citep{hur2013esophageal}} & \fixedtablecell{\raggedright}{Cancer (2013)} & \fixedtablecell{\raggedleft}{510} & \fixedtablecell{\raggedleft}{33} & \fixedtablecell{\raggedleft}{52} & \stackedoutcomes{40}{76.9}{3}{5.8}{9}{17.3} \\
\rowcolor{black!4}
\fixedtablecell{\raggedright}{Second cancers after young-onset colorectal cancer~\citep{he2018second}} & \fixedtablecell{\raggedright}{Cancer Med. (2018)} & \fixedtablecell{\raggedleft}{43} & \fixedtablecell{\raggedleft}{59} & \fixedtablecell{\raggedleft}{124} & \stackedoutcomes{95}{76.6}{14}{11.3}{15}{12.1} \\
\fixedtablecell{\raggedright}{Lung cancer incidence by sex, race, and histology~\citep{meza2015lung}} & \fixedtablecell{\raggedright}{PLOS ONE (2015)} & \fixedtablecell{\raggedleft}{368} & \fixedtablecell{\raggedleft}{36} & \fixedtablecell{\raggedleft}{111} & \stackedoutcomes{80}{72.1}{7}{6.3}{24}{21.6} \\
\rowcolor{black!4}
\fixedtablecell{\raggedright}{Age disparities in colorectal cancer incidence~\citep{bailey2015colorectal}} & \fixedtablecell{\raggedright}{JAMA Surg. (2015)} & \fixedtablecell{\raggedleft}{1,102} & \fixedtablecell{\raggedleft}{27} & \fixedtablecell{\raggedleft}{82} & \stackedoutcomes{59}{72.0}{7}{8.5}{16}{19.5} \\
\fixedtablecell{\raggedright}{Fatal prostate cancer incidence by race~\citep{kelly2017fatal}} & \fixedtablecell{\raggedright}{Eur. Urol. (2017)} & \fixedtablecell{\raggedleft}{111} & \fixedtablecell{\raggedleft}{32} & \fixedtablecell{\raggedleft}{63} & \stackedoutcomes{45}{71.4}{2}{3.2}{16}{25.4} \\
\rowcolor{black!4}
\fixedtablecell{\raggedright}{Intracranial glioma incidence and mortality~\citep{lin2021glioma}} & \fixedtablecell{\raggedright}{Front. Oncol. (2021)} & \fixedtablecell{\raggedleft}{116} & \fixedtablecell{\raggedleft}{67} & \fixedtablecell{\raggedleft}{94} & \stackedoutcomes{66}{70.2}{2}{2.1}{26}{27.7} \\
\fixedtablecell{\raggedright}{Long-term mortality in young cancer survivors~\citep{moke2021survivors}} & \fixedtablecell{\raggedright}{Cancers (2021)} & \fixedtablecell{\raggedleft}{23} & \fixedtablecell{\raggedleft}{62} & \fixedtablecell{\raggedleft}{178} & \stackedoutcomes{123}{69.1}{5}{2.8}{50}{28.1} \\
\rowcolor{black!4}
\fixedtablecell{\raggedright}{Breast cancer incidence by age and tumor features~\citep{jemal2007breast}} & \fixedtablecell{\raggedright}{Breast Cancer Res. (2007)} & \fixedtablecell{\raggedleft}{352} & \fixedtablecell{\raggedleft}{31} & \fixedtablecell{\raggedleft}{67} & \stackedoutcomes{39}{58.2}{3}{4.5}{25}{37.3} \\
\fixedtablecell{\raggedright}{Esophageal versus gastric cardia adenocarcinoma~\citep{elserag2002esophageal}} & \fixedtablecell{\raggedright}{Gut (2002)} & \fixedtablecell{\raggedleft}{266} & \fixedtablecell{\raggedleft}{34} & \fixedtablecell{\raggedleft}{51} & \stackedoutcomes{29}{56.9}{2}{3.9}{20}{39.2} \\
\rowcolor{black!4}
\fixedtablecell{\raggedright}{Metastatic cancer incidence and survival~\citep{hudock2023metastatic}} & \fixedtablecell{\raggedright}{Commun. Med. (2023)} & \fixedtablecell{\raggedleft}{68} & \fixedtablecell{\raggedleft}{51} & \fixedtablecell{\raggedleft}{96} & \stackedoutcomes{52}{54.2}{4}{4.2}{40}{41.7} \\
\fixedtablecell{\raggedright}{Cervical cancer incidence by race and histology~\citep{wang2004cervical}} & \fixedtablecell{\raggedright}{Cancer (2004)} & \fixedtablecell{\raggedleft}{448} & \fixedtablecell{\raggedleft}{46} & \fixedtablecell{\raggedleft}{120} & \stackedoutcomes{64}{53.3}{1}{0.8}{55}{45.8} \\
\midrule
\multicolumn{1}{@{}l}{\textbf{Total}} & & & \multicolumn{1}{r}{\textbf{901}} & \multicolumn{1}{r}{\textbf{2,190}} & \multicolumn{3}{r@{}}{\makebox[\outcomespan][s]{{\setlength{\fboxsep}{1pt}\colorbox{reprobar}{\textbf{1,694 (77.4\%)}}}\hfil{\setlength{\fboxsep}{1pt}\colorbox{closebar}{\textbf{76 (3.5\%)}}}\hfil{\setlength{\fboxsep}{1pt}\colorbox{unreprobar}{\textbf{420 (19.2\%)}}}}} \\
\bottomrule
\end{tabularx}

\end{table}

\subsection{\benchmarkname{} Construction Challenges}
\label{sec:reproduction-challenges}

We apply our pipeline to published SEER findings. Reconstructing these studies presents two challenges:

\textbf{Historical data availability.} Detroit was excluded from the April 2022 release, changing SEER 9, 13, and 18 to SEER 8, 12, and 17 \citep{seerdatachanges}. We focus on SEER 9 studies because a retained supplement covers 1975--2018, including Detroit \citep{seersupplemental2020}. Current releases cannot recover the exact historical SEER 13 and 18 cohorts. Appendix~\ref{app:seer-evolution} details remaining limitations.

\textbf{Fragmented data and documentation.}
Epidemiological studies often rely on external population data and relevant manuals. We curate these resources manually through an iterative process with expert input. Missing inputs often become apparent only as individual analyses are reconstructed, and identifying what is missing can itself require tracing the analysis step by step. For example, reproducing an incidence rate requires identifying an appropriate population denominator, locating its source, and checking that its geographic coverage, calendar years, and demographic definitions match the cancer records. Selecting among candidate sources also requires assessing their suitability and consulting the relevant manuals. This process therefore requires manual curation and expert judgment, rather than simply retrieving a known list of files. 

We curated SEER 9 case table, four county-population tables, 34 SEER*Explorer tables, and associated documentation on coding, staging, registry mappings, and population methods (Appendix~\ref{app:seer-resources}). These 39 tables contain a combined 713 columns and 198,255,448 rows (Table~\ref{tab:seer-table-inventory}). The accompanying 40 documentation resources include 11 PDF manuals and reference documents totaling 816 pages, 28 web pages, and one variable-dictionary spreadsheet (Table~\ref{tab:seer-manual-inventory}). These data tables and documentation resources form the execution environment for \benchmarkname{}.

\subsection{\benchmarkname{} Benchmark and Reproduction Outcome}
\begin{table}[!t]
\centering
\caption{Classification of all 420 non-reproduced atomic claims from 20 papers in \benchmarkname{}.}\label{tab:reproduction-issues}
\footnotesize
\setlength{\tabcolsep}{3pt}
\begin{tabularx}{\linewidth}{@{}>{\raggedright\arraybackslash}p{0.23\linewidth}>{\raggedright\arraybackslash}Xrr@{}}
\toprule
Category & Meaning & Claims & \% \\
\midrule
Data issues & Unavailable data (212 cases) or a data-related explanation such as mismatches attributed to revised records and population denominators (39 cases). & 251 & 59.8 \\
\addlinespace[2pt]
Insufficiently specified analysis & The available paper and documentation do not uniquely specify the target analysis. & 47 & 11.2 \\
\addlinespace[2pt]
Unavailable-software approximations & Method substitutions involving approximations of unavailable software such as NCI Joinpoint. & 28 & 6.7 \\
\addlinespace[2pt]
Agent-side errors & Departures from the reported method or comparison (25), incorrect cohort (13), statistical inference (3), denominator or standardization (6), or an omitted analysis (1). & 48 & 11.4 \\
\addlinespace[2pt]
Suspected evaluation errors & Suspected errors in our claim extraction or grading. & 20 & 4.8 \\
\addlinespace[2pt]
Unresolved cause & The evidence does not identify a sufficiently specific cause of non-reproduction. & 26 & 6.2 \\
\bottomrule
\end{tabularx}
\end{table}

We collect the ten most cited eligible papers and ten more recent papers in \benchmarkname{}. Using GPT-5.6 Sol xhigh via Codex, our reproduction agent reproduces 1,694 of 2,190 atomic claims (77.4\%); 76 are close (3.5\%) and 420 are not reproduced (19.2\%; Table~\ref{tab:seer-twenty-reproduction}). We classify the 420 failures using Prompt~\ref{prompt:reproduction-failure-classification} and review the diagnoses against saved evidence. Data issues account for 251 (59.8\%), insufficiently specified analyses for 47 (11.2\%), unavailable-software approximations for 28 (6.7\%), and agent-side errors for 48 (11.4\%; Table~\ref{tab:reproduction-issues}).

The benchmark packages 39 data tables and 40 documentation resources (Appendix~\ref{app:seer-resources}), together with the 2,190 atomic claims over 20 papers. Each claim $a_{ij}\in A_i$ links to its paper $p$, parent finding $f_i$, verdict $v_{ij}$, and reproduction evidence $e_i$, including code and available execution outputs.

\subsection{Robustness Analysis of \benchmarkname{}}
\label{sec:robustness-results}

\subsubsection{Independent Report Generation}

For each paper, we independently ask Opus 4.8 max via Claude Code and GPT-5.6 Sol xhigh via Codex to generate a report. GPT-5.6 Sol then compares each of the 40 reports with its corresponding paper for contradictions on their shared research question (Prompt~\ref{prompt:conclusion-comparison}). The judge identifies no contradictions in the central conclusions of the 39 comparable reports (19 Sol and 20 Opus). One Sol report analyzes a nonoverlapping study period and is not comparable. It marked 3 Opus reports as containing contradictions in supporting statements, but these do not overturn their central conclusions upon closer scrutiny. These results suggest that independent analyses generally reach central conclusions non-contradictory with the published papers.

\subsubsection{Perturbation Agent and Direct Coding Agent Analysis}

\begin{table}[!t]
\centering
\caption{Model-assessed consequential links (scores 7--9): finding counts by paper and analyst. Counts include individual links and members of consequential joint links, deduplicated per analyst.}
\label{tab:finding-sensitivity-coverage}
\scriptsize
\setlength{\tabcolsep}{2.5pt}
\renewcommand{\arraystretch}{1.02}
\begin{tabularx}{\linewidth}{@{}>{\raggedright\arraybackslash}X r rrrr rr >{\columncolor{black!5}}r >{\columncolor{black!5}}r@{}}
\toprule
\multirow{3}{*}{Paper} & \multirow{3}{*}{\shortstack{Eligible\\findings}} & \multicolumn{8}{c}{Findings with consequential links} \\
\cmidrule(lr){3-10}
 & & \multicolumn{4}{c}{Perturbation Agent} & \multicolumn{2}{c}{Direct coding agent} & \multicolumn{2}{c}{\cellcolor{black!5}Found by} \\
\cmidrule(lr){3-6}\cmidrule(lr){7-8}\cmidrule(lr){9-10}
 & & Sol & Opus & Kimi & GLM & Codex & \shortstack{Claude\\Code} & \shortstack{$\geq 1$\\analyst} & \shortstack{$\geq 2$\\analysts} \\
\midrule
Bone sarcoma survival by tumor site & 55 & 2 & \textendash & 2 & \textendash & \textendash & \textendash & 4 & \textendash \\
Sex disparities in cancer incidence & 31 & \textendash & \textendash & \textendash & \textendash & \textendash & \textendash & \textendash & \textendash \\
Inflammatory breast cancer incidence and survival & 31 & \textendash & 3 & \textendash & \textendash & 2 & 3 & 4 & 3 \\
Thyroid cancer trends & 24 & \textendash & \textendash & \textendash & \textendash & 2 & \textendash & 2 & \textendash \\
Age at oropharyngeal cancer diagnosis & 53 & \textendash & 3 & \textendash & \textendash & \textendash & \textendash & 3 & \textendash \\
Adenoid cystic carcinoma of the breast & 37 & \textendash & \textendash & \textendash & \textendash & 4 & \textendash & 4 & \textendash \\
Adult diffuse glioma incidence & 40 & \textendash & \textendash & \textendash & \textendash & 2 & \textendash & 2 & \textendash \\
Childhood cancer incidence and mortality & 28 & \textendash & \textendash & \textendash & 1 & \textendash & \textendash & 1 & \textendash \\
Head and neck mucosal melanoma incidence & 12 & \textendash & \textendash & \textendash & 1 & 2 & 1 & 2 & 2 \\
Esophageal adenocarcinoma incidence and mortality & 23 & \textendash & \textendash & \textendash & 2 & \textendash & \textendash & 2 & \textendash \\
Second cancers after young-onset colorectal cancer & 44 & \textendash & \textendash & \textendash & \textendash & 1 & 2 & 2 & 1 \\
Lung cancer incidence by sex, race, and histology & 13 & 1 & \textendash & \textendash & \textendash & \textendash & \textendash & 1 & \textendash \\
Age disparities in colorectal cancer incidence & 10 & \textendash & \textendash & \textendash & 1 & \textendash & \textendash & 1 & \textendash \\
Fatal prostate cancer incidence by race & 20 & \textendash & \textendash & 5 & \textendash & 3 & 3 & 6 & 3 \\
Intracranial glioma incidence and mortality & 47 & \textendash & \textendash & 1 & \textendash & \textendash & \textendash & 1 & \textendash \\
Long-term mortality in young cancer survivors & 40 & 2 & 4 & 6 & \textendash & 2 & \textendash & 6 & 4 \\
Breast cancer incidence by age and tumor features & 15 & \textendash & 2 & \textendash & \textendash & \textendash & \textendash & 2 & \textendash \\
Esophageal versus gastric cardia adenocarcinoma & 15 & \textendash & \textendash & \textendash & \textendash & 5 & \textendash & 5 & \textendash \\
Metastatic cancer incidence and survival & 22 & \textendash & 1 & 2 & \textendash & 1 & 3 & 4 & 1 \\
Cervical cancer incidence by race and histology & 19 & \textendash & \textendash & 1 & \textendash & 2 & \textendash & 3 & \textendash \\
\midrule
\textbf{Total} & \textbf{579} & \textbf{5} & \textbf{13} & \textbf{17} & \textbf{5} & \textbf{26} & \textbf{12} & \textbf{55} & \textbf{14} \\
\bottomrule
\end{tabularx}
\end{table}

We run the Perturbation Agent with GPT-5.6 Sol \citep{openai2026gpt56sol}, Opus 4.8 \citep{anthropic2026opus48}, Kimi K3 \citep{kimi2026k3}, and GLM 5.3 \citep{zai2026glm53}, together with direct coding agent analyses through GPT-5.6 Sol xhigh via Codex and Opus 4.8 max via Claude Code. These correspond to methods B and C in Figure~\ref{fig:analysis-pipeline}. GPT-5.6 Sol then assigned scores of 7--9 to 54 results (Prompt~\ref{prompt:consequentiality-ordinal-v3}). These scores indicate a model-assessed major qualification, loss of support, or reversal of a principal conclusion and help identify shortlisted candidates for expert examination.

We use GPT-5.6 Sol to link these high-score results back to the original findings in each paper (Prompt~\ref{prompt:finding-consequence-mapping}). A model-assessed consequential link requires reported evidence of a reversal, loss of support, major qualification, or material change to a quantity central to one specific finding or a conclusion that the paper draws jointly from multiple findings. In the latter case, we link the analysis to each supporting finding. Table~\ref{tab:finding-sensitivity-coverage} shows that 55 of the 579 fully reproduced findings (9.5\%) have such a link from at least one analyst, and 14 (2.4\%) from at least two analysts.

\begin{wrapfigure}[15]{r}{0.48\textwidth}
\captionsetup{skip=4pt}
\centering
\includegraphics[width=\linewidth]{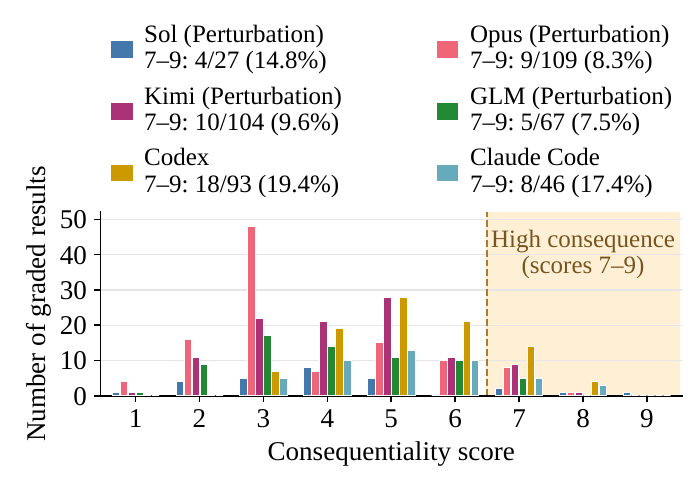}
\caption{GPT-5.6 Sol consequence scores for analyses of published papers; shading marks scores 7--9.}
\label{fig:human-score-distribution}
\end{wrapfigure}

\textbf{Discussion}: As mentioned in Section~\ref{sec:analysis}, we restrict this analysis to fully reproduced findings. This provides executable baselines whose results passed the reproduction criteria in Section~\ref{sec:reproduction}. The analysts can therefore compare an alternative against a reconstructed result supported by reproduction evidence. We further consider two questions:

\textbf{Do different analysts complement each other?} Results in Table~\ref{tab:finding-sensitivity-coverage} show that most of the affected findings---41 of 55 (74.5\%)---are identified by only one analyst. This suggests that different analysts identify largely complementary sets of potentially consequential analytical choices.

\textbf{Distribution of consequence scores.} Among results with a 1--9 grade (Figure~\ref{fig:human-score-distribution}), scores 7--9 account for 7.5--14.8\% of Perturbation Agent comparisons, 19.4\% of Codex results, and 17.4\% of Claude Code results. Thus, most graded results do not reach the high-consequence range.

\begin{table}[!t]
\centering
\caption{Three expert-reviewed cases, one per assessment category.}\label{tab:direct-change-examples}
\scriptsize
\setlength{\tabcolsep}{3pt}
\renewcommand{\arraystretch}{0.98}
\begin{tabularx}{\linewidth}{@{}>{\raggedright\arraybackslash}p{0.29\linewidth}>{\raggedright\arraybackslash}p{0.31\linewidth}>{\raggedright\arraybackslash}X@{}}
\toprule
Paper and original conclusion & Alternative analysis and result & Expert assessment \\
\midrule
\textbf{Bone sarcoma survival by tumor site.} Sol, Perturbation Agent; score 9. Adjusted cancer-specific mortality did not significantly improve in the 2000s or 2010s relative to the 1990s. & Replace the SEER ``Bones and Joints'' recode with direct primary-site codes C40.0--C41.9. The implementation added 3,974 records; both period comparisons then showed significant improvements. & \textbf{Likely less defensible alternative.} The added records change the disease population, and closer examination of the database contents and manuals suggests that the alternative is less defensible than the original choice. \\
\midrule
\textbf{Adenoid cystic carcinoma of the breast.} Codex, direct analysis; score 8. Nonsignificant standardized incidence ratios (SIRs) were interpreted as an absence of associated cancers. & Interpret the estimates as no statistically detected excess, while acknowledging that the confidence intervals still permit increased risk. Failure to reject the null does not establish absence of excess risk. & \textbf{Limited qualification.} The distinction is reasonable, but the paper already qualifies its conclusions with caveats. The reviewer judged this a clarification of interpretation rather than a conclusion-changing result. \\
\midrule
\textbf{Head and neck mucosal melanoma incidence.} Claude Code, direct analysis; score 8. Increasing incidence was attributed primarily to nasal-cavity disease. & Directly compare nasal and non-nasal trends rather than infer a difference from their separate significance tests. The analyst reported positive trends in both groups and a nonsignificant interaction ($p=0.46$). & \textbf{Flagged for further review.} The comparison may identify an overstatement, but noisy data and model-fit concerns complicate interpretation. Closer manuscript review and statistical consultation are needed; a nonsignificant interaction does not establish equal trends. \\
\bottomrule
\end{tabularx}
\end{table}

\subsubsection{Expert Analysis}

One coauthor with medical expertise and familiarity with SEER data manually examined 10 cases scored 7--9 in depth (five each from Perturbation Agent and Direct Coding Agent Analysis). Among the 5 cases from the Perturbation Agent, 2 suggested consequential changes, but closer expert scrutiny indicated that the alternatives were likely less defensible than the original analytical choices; the remaining 3 offered only limited qualifications to the papers' conclusions. Among the 5 cases from direct coding agent analysis, 4 likewise offered only limited qualifications. The remaining 1 case identified a potentially overstated comparison between nasal and non-nasal incidence trends, but determining its consequence requires closer manuscript review and statistical consultation. Thus, 9 cases did not establish a consequential change under a defensible alternative, while 1 remains flagged for further review. Table~\ref{tab:direct-change-examples} illustrates 3 cases with one from each category: a likely less defensible alternative, a limited qualification to the paper's conclusions, and a potentially consequential issue requiring further review.

Overall, our pipeline narrows the material requiring expert attention by autonomously reconstructing published findings, executing alternative analyses, and prioritizing potentially consequential results. Across 579 fully reproduced findings, it identifies 55 with model-assessed consequential links, providing a focused shortlist and supporting evidence for expert review. Experts can therefore concentrate on whether the flagged alternatives are scientifically defensible and will materially change the conclusions, rather than reconstructing and testing each analysis from scratch manually.

\section{Conclusion}

We introduced \benchmarkname{}, a benchmark of 20 published SEER studies, and an AI-assisted pipeline for evaluating reproducibility and robustness to analytical choices. Our reproduction agent recovered 77.4\% of atomic claims. Automated robustness assessments then prioritized 55 of 579 fully reproduced findings (9.5\%) for expert review, providing a focused shortlist with supporting analytical evidence. Expert examination of 10 flagged cases clarified why 9 did not establish consequential changes under defensible alternatives and identified 1 potentially consequential issue requiring further investigation. These results illustrate how automated analysis can focus expert attention on questions that warrant closer scientific scrutiny.

SEER provides an early, high-stakes testbed for this approach. Longer term, such checks could become part of scientific infrastructure: published findings could be routinely reassessed as data, methods, and analytical alternatives evolve, with potentially consequential sensitivities prioritized for expert review.

\newpage
\subsection*{AI Use Statement}

\textbf{Uses of generative AI.} In this work, we used generative AI tools to help develop theoretical models or conceptual frameworks; propose or refine hypotheses; design or provide feedback on research methodology or experiments; implement methods; clean and reformat datasets; support qualitative and thematic data analysis; interpret results; create or modify scientific figures or images; create or edit software code; draft parts of the paper; create artifacts; summarize or analyze existing literature; discover research topics or identify gaps; brainstorm; source and search for information; improve the paper's readability; identify relevant literature; and format references.

\textbf{Activities without generative AI.} We did not use generative AI tools to suggest experimental parameters, suggest a structure for the paper, or propose a title or keywords.

\textbf{Activities not applicable to this work.} Generating synthetic datasets, formulating mathematical claims, providing critical ingredients for mathematical proofs, assisting with proof writing, translating, formulating survey or interview questions, and transcribing research recordings are not applicable to this work.

\textbf{Additional disclosure and author responsibility.}
We have reviewed all AI-assisted work. For example, the final writing has been double-checked by authors. We have reviewed all relevant literature manually found by LLMs. We take responsibility for the final content of this work, including text, claims or artifacts produced with the aid of generative AI.

\subsection*{Ethics Statement}

This work aims to support scientific reliability by evaluating the reproducibility and analytical robustness of published cancer epidemiology findings. We acknowledge that errors in our pipeline, including but not limited to data selection, LLM-generated code, interpretation, and grading, may lead to incorrect assessments. Failure to reproduce a finding may also reflect unavailable historical data (as documented in Section~\ref{sec:reproduction-challenges}). Sensitivity to an alternative analysis does not invalidate the original study and requires careful expert judgment to assess its significance (as discussed in Section~\ref{sec:results}). Our goal is to help experts identify findings that warrant closer examination, making rigorous scientific review more accessible while keeping expert judgment central to the process.

\subsection*{Reproducibility Statement}

We document the SEER data releases and registry coverage in Appendix~\ref{app:seer-evolution}, and the database resources, manuals, source links, and file formats in Appendix~\ref{app:seer-resources}. Table~\ref{tab:seer-twenty-reproduction} identifies the exact 20 papers evaluated in \benchmarkname{}. Appendix~\ref{app:llm-research-reports} provides the research prompts and report-generation settings, while Appendix~\ref{app:analysis-implementation} describes the analysis implementation and Appendix~\ref{app:prompts} provides the LLM prompts. Section~\ref{sec:results} identifies the tested analysis models (GPT-5.6 Sol, Opus 4.8, GLM 5.3, and Kimi K3) and the coding-agent configurations used in the comparison experiments.

\subsection*{Acknowledgment}

We thank Ivan Zhou, Qing (Veronica) Lyu, Xinyi (Cindy) Wang, Muhammad Shahir Abdurrahman, and Xing Chen at Databricks; and Harshit Joshi, Zikai (Cyrus) Zhou, Yanzhe Zhang, and Yuchen Jiang at Stanford for discussions that shaped the project.

Shicheng Liu’s academic research outside Databricks was supported in part by the Verdant Foundation, the Hasso Plattner Institute, Itaú Unibanco, BMO Financial Group, and the Stanford Human-Centered Artificial Intelligence (HAI) Institute. We acknowledge the National Artificial Intelligence Research Resource (NAIRR) Pilot and Microsoft Azure for contributing to the results in this work.

\clearpage
\bibliography{iclr2027_conference}
\bibliographystyle{iclr2027_conference}

\clearpage
\appendix
\etocdepthtag.toc{appendix}
\begingroup
\etocsettagdepth{main}{none}
\etocsettagdepth{appendix}{subsection}
\etocsettocstyle{\section*{Appendix}}{}
\tableofcontents
\endgroup
\clearpage

\section{SEER Data Evolution and Study Scope}
\label{app:seer-evolution}

\begin{figure}[htbp]
\centering
\includegraphics[width=\linewidth]{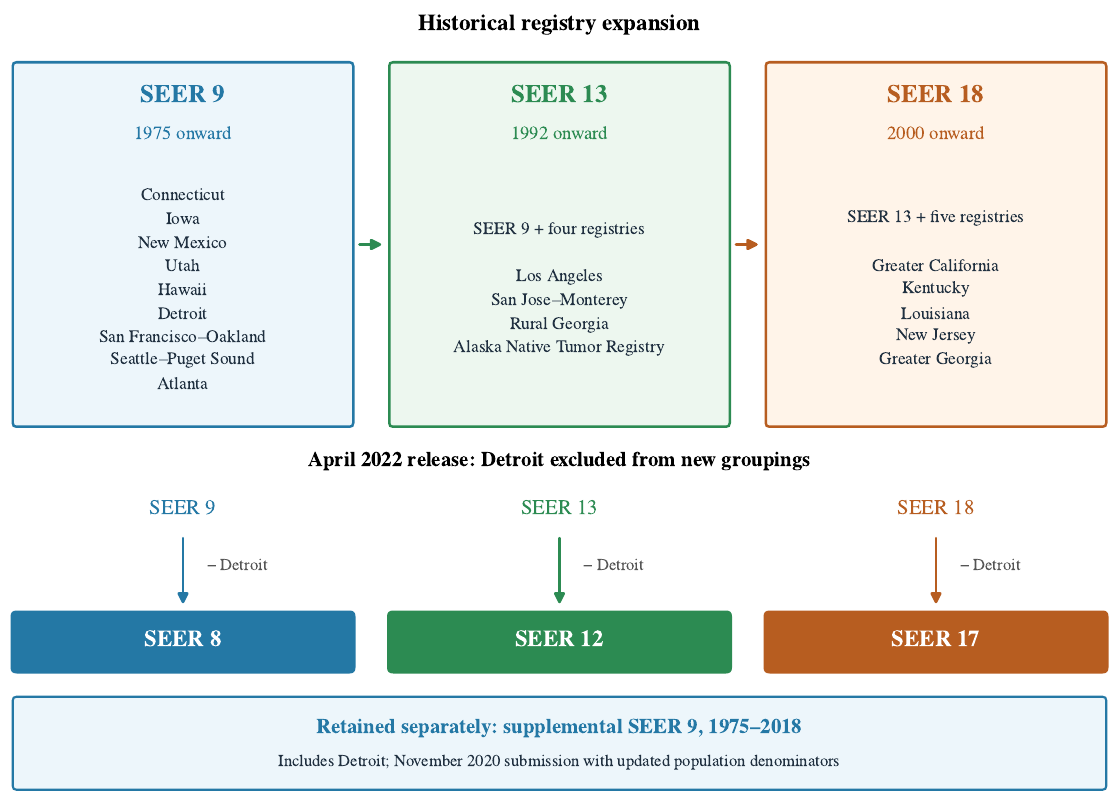}
\caption{Evolution of the SEER registry groupings relevant to our study. Historical SEER 13 and 18 expand the SEER 9 cohort; their successor groupings exclude Detroit. The separately retained SEER 9 supplement covers 1975--2018. Registry memberships and retained coverage follow NCI documentation \citep{seerregistrygroupings,seersupplemental2020}.}
\label{fig:seer-data-evolution}
\end{figure}

\paragraph{Registry Overview.}
SEER 9, SEER 13, and SEER 18 identify different geographic cohorts, with standard diagnosis-year coverage beginning in 1975, 1992, and 2000, respectively. Figure~\ref{fig:seer-data-evolution} summarizes their expansion and subsequent changes. With the November 2021 submission, released in April 2022, Detroit was excluded from the new databases, yielding SEER 8, 12, and 17. Massachusetts was subsequently excluded from the November 2024 submission, changing SEER 22 to SEER 21 \citep{seerdatachanges,seerregistrygroupings}.

\paragraph{Why we focus on SEER 9?}
The supplemental November 2020 database preserves SEER 9 records, including Detroit, through 2018. We use this retained cohort and select papers whose analyses do not require additional registries. It is in general difficult to reconstruct historical SEER 13 and SEER 18 datasets using the newer, Detroit-excluding releases (SEER 12 and SEER 17), since each case does not record the exact associated registry.

\paragraph{Release and variable differences remain.}
Retaining the registry grouping does not guarantee identical historical inputs. For example, the November 2020 submission changed the default malignant-behavior definition and excluded approximately 10,000 previously included, no-longer-reportable borderline cases, predominantly ovarian tumors \citep{seerbehaviorchanges}. Our local failure diagnostics also identify unavailable radiation-treatment fields, preventing reconstruction of some treatment proportions even when registry coverage matches. We retain these limitations when interpreting claim-level reproduction outcomes. For instance, these cases would count towards ``Data Issues'' in Table~\ref{tab:reproduction-issues}.

\clearpage
\section{Supplementary SEER Tables and Manuals}
\label{app:seer-resources}

The curated \benchmarkname{} environment contains the SEER 9 case table, 4 population denominator tables, 34 SEER*Explorer aggregate tables, and 40 documentation files or reference pages.

\paragraph{Population denominators.}
The four population tables provide county-level annual estimates by age, sex, and race, with an expanded race/Hispanic-origin scheme available from 1990. Each series is supplied both in 20 age groups and in single-year ages through 89 with a 90+ category (Table~\ref{tab:seer-table-inventory}). These are all-state files: rates for a SEER cohort require restricting them to the corresponding registry counties \citep{seerpopulationdata}.

\paragraph{SEER*Explorer aggregates.}
We include 34 workbooks from the official November 2024 SEER*Explorer archive as queryable tables. Table~\ref{tab:seer-table-inventory} lists each table with its column count, row count, and description. They supplement case-level computation with aggregated statistics.

\paragraph{Manuals and reference definitions.}
Table~\ref{tab:seer-manual-inventory} lists the 40 documentation resources supplied in \benchmarkname{}: 28 HTML pages, 11 PDFs, and one XLSX spreadsheet. The collection covers SEER*Stat dictionaries, population denominators and standardization, coding and staging, variable recodes, and registry coverage. It includes both recent manuals and a historical-manual index.

\begingroup
\footnotesize
\setlength{\tabcolsep}{3pt}
\renewcommand{\arraystretch}{1.1}
\begin{longtable}{@{}>{\raggedright\arraybackslash}p{0.36\linewidth}>{\raggedright\arraybackslash}p{0.53\linewidth}l@{}}
\caption{Documentation resources provided in \benchmarkname{}. Source URLs are those recorded for the local copies; formats were verified from the files.}\label{tab:seer-manual-inventory}\\
\toprule
Resource & Source URL & Format \\
\midrule
\endfirsthead
\multicolumn{3}{l}{\tablename~\thetable{} (continued)}\\
\toprule
Resource & Source URL & Format \\
\midrule
\endhead
\midrule
\multicolumn{3}{r}{Continued on next page}\\
\endfoot
\bottomrule
\endlastfoot
\multicolumn{3}{@{}l}{\textbf{SEER*Stat resources (6)}}\\
\addlinespace[2pt]
SEER*Stat November 2025 submission overview & {\scriptsize\url{https://seer.cancer.gov/data-software/documentation/seerstat/nov2025/}} & HTML \\
Variable dictionary spreadsheet for SEER*Stat variables & {\scriptsize\url{https://seer.cancer.gov/data-software/documentation/seerstat/nov2025/seerstat-variable-dictionary-nov2025.xlsx}} & XLSX \\
Variable dictionary PDF for SEER*Stat variables & {\scriptsize\url{https://seer.cancer.gov/data-software/documentation/seerstat/nov2025/seerstat-variable-dictionary-nov2025.pdf}} & PDF \\
Text data file description for SEER Research and Research Plus & {\scriptsize\url{https://seer.cancer.gov/data-software/documentation/seerstat/nov2025/TextData.FileDescription-nov2025.pdf}} & PDF \\
SEER product comparison and availability notes & {\scriptsize\url{https://seer.cancer.gov/data/product-comparison.html}} & HTML \\
SEER data documentation hub & {\scriptsize\url{https://seer.cancer.gov/data-software/documentation/}} & HTML \\
\multicolumn{3}{@{}l}{\textbf{Population resources (10)}}\\
\addlinespace[2pt]
U.S. county population data overview & {\scriptsize\url{https://seer.cancer.gov/popdata/}} & HTML \\
Population data download page and file layout links & {\scriptsize\url{https://seer.cancer.gov/popdata/download.html}} & HTML \\
Population data dictionary for fixed-width fields & {\scriptsize\url{https://seer.cancer.gov/popdata/popdic.html}} & HTML \\
Single-year-age population estimate methodology & {\scriptsize\url{https://seer.cancer.gov/popdata/singleages.html}} & HTML \\
NCI modifications to county population estimates & {\scriptsize\url{https://seer.cancer.gov/popdata/modifications.html}} & HTML \\
Katrina/Rita population adjustment notes & {\scriptsize\url{https://seer.cancer.gov/popdata/hurricane\_adj.html}} & HTML \\
Woods \& Poole technical documentation linked by SEER & {\scriptsize\url{https://seer.cancer.gov/popdata/W\%26P\_ICen\_Tech\_Documentation.pdf}} & PDF \\
U.S. populations hub & {\scriptsize\url{https://seer.cancer.gov/data-software/uspopulations.html}} & HTML \\
Standard population data hub & {\scriptsize\url{https://seer.cancer.gov/stdpopulations/}} & HTML \\
Use of the 2000 U.S. standard population & {\scriptsize\url{https://seer.cancer.gov/stdpopulations/2000stdpop-use.html}} & HTML \\
\multicolumn{3}{@{}l}{\textbf{Coding and staging (10)}}\\
\addlinespace[2pt]
SEER coding and staging manual page & {\scriptsize\url{https://seer.cancer.gov/tools/codingmanuals/}} & HTML \\
Historical coding manuals index & {\scriptsize\url{https://seer.cancer.gov/tools/codingmanuals/historical.html}} & HTML \\
2025 SEER Program Coding and Staging Manual & {\scriptsize\url{https://seer.cancer.gov/manuals/2025/SPCSM\_2025\_MainDoc.pdf}} & PDF \\
2025 county codes & {\scriptsize\url{https://seer.cancer.gov/manuals/2025/SPSCM\_2025\_Appendix\_A.pdf}} & PDF \\
2025 country and state codes & {\scriptsize\url{https://seer.cancer.gov/manuals/2025/SPCSM\_2025\_Appendix\_B.pdf}} & PDF \\
2025 race and nationality descriptions & {\scriptsize\url{https://seer.cancer.gov/manuals/2025/SPCSM\_2025\_Appendix\_D.pdf}} & PDF \\
2025 reportable and non-reportable examples & {\scriptsize\url{https://seer.cancer.gov/manuals/2025/SPCSM\_2025\_Appendix\_E.pdf}} & PDF \\
2025 summary of changes & {\scriptsize\url{https://seer.cancer.gov/manuals/2025/SPCSM\_2025\_Changelog.pdf}} & PDF \\
2026 SEER Program Coding and Staging Manual & {\scriptsize\url{https://seer.cancer.gov/manuals/2026/SPCSM\_2026\_MainDoc.pdf}} & PDF \\
2026 county codes & {\scriptsize\url{https://seer.cancer.gov/manuals/2026/SPSCM\_2026\_Appendix\_A.pdf}} & PDF \\
\multicolumn{3}{@{}l}{\textbf{Variable and recode definitions (10)}}\\
\addlinespace[2pt]
Variable and recode definitions hub & {\scriptsize\url{https://seer.cancer.gov/analysis/}} & HTML \\
Site recode documentation & {\scriptsize\url{https://seer.cancer.gov/siterecode/}} & HTML \\
Incidence site recode variables & {\scriptsize\url{https://seer.cancer.gov/analysis/incidence.html}} & HTML \\
Cause of death recode & {\scriptsize\url{https://seer.cancer.gov/codrecode/}} & HTML \\
Cause-specific death classification & {\scriptsize\url{https://seer.cancer.gov/causespecific/}} & HTML \\
Race and Hispanic ethnicity definitions & {\scriptsize\url{https://seer.cancer.gov/seerstat/variables/seer/race\_ethnicity/}} & HTML \\
Cancer stage variable overview & {\scriptsize\url{https://seer.cancer.gov/analysis/stage.html}} & HTML \\
Summary Stage 2018 documentation & {\scriptsize\url{https://seer.cancer.gov/tools/ssm/}} & HTML \\
Months survived based on complete dates & {\scriptsize\url{https://seer.cancer.gov/survivaltime/}} & HTML \\
Tumor size over time recode & {\scriptsize\url{https://seer.cancer.gov/tumorsize/}} & HTML \\
\multicolumn{3}{@{}l}{\textbf{Registry coverage (4)}}\\
\addlinespace[2pt]
SEER registries overview & {\scriptsize\url{https://seer.cancer.gov/registries/}} & HTML \\
List of SEER registries & {\scriptsize\url{https://seer.cancer.gov/registries/list.html}} & HTML \\
Registry groupings used in SEER data and statistics & {\scriptsize\url{https://seer.cancer.gov/registries/terms.html}} & HTML \\
Population characteristics of SEER registry areas & {\scriptsize\url{https://seer.cancer.gov/registries/characteristics.html}} & HTML \\
\end{longtable}
\endgroup

\begingroup
\footnotesize
\setlength{\tabcolsep}{3pt}
\renewcommand{\arraystretch}{1.0}
\begin{longtable}{@{}>{\raggedright\arraybackslash}p{0.40\linewidth}r r >{\raggedright\arraybackslash}p{0.34\linewidth}@{}}
\caption{The 39 curated tables, with schema column counts and loaded row counts. SEER*Explorer tables use the November 2024 archive; RUCC denotes rural--urban continuum codes.}\label{tab:seer-table-inventory}\\
\toprule
Table & Cols. & Rows & Description \\
\midrule
\endfirsthead
\multicolumn{4}{l}{\tablename~\thetable{} (continued)}\\
\toprule
Table & Cols. & Rows & Description \\
\midrule
\endhead
\midrule
\multicolumn{4}{r}{Continued on next page}\\
\endfoot
\bottomrule
\endlastfoot
SEER 9 case records & 250 & 5,075,266 & Case-level cancer-registry export from the SEER 9 Program. \\
Population 1969--2024, 20 age groups & 13 & 17,485,723 & County denominators; White/Black/Other. \\
Population 1990--2024, 20 age groups & 13 & 22,178,259 & County denominators; expanded race/Hispanic origin. \\
Population 1969--2024, single-year ages & 13 & 69,655,933 & County denominators; White/Black/Other. \\
Population 1990--2024, single-year ages & 13 & 78,667,287 & County denominators; expanded race/Hispanic origin. \\
Recent trends annual SEER incidence rates & 13 & 1,648,320 & Annual incidence rates. \\
Recent trends SEER incidence trend estimates & 17 & 104,360 & Incidence trend estimates. \\
5-year age adjusted incidence rates & 11 & 67,360 & Five-year age-adjusted incidence rates. \\
Long-term trends annual SEER incidence rates & 12 & 488,160 & Annual incidence rates. \\
Long-term trends SEER incidence trend estimates & 16 & 38,815 & Incidence trend estimates. \\
Incidence rates by age at diagnosis & 12 & 59,360 & Incidence rates stratified by age at diagnosis. \\
Stage distribution & 8 & 51,440 & Stage distribution. \\
Median age at diagnosis & 6 & 1,592 & Median age at diagnosis estimates. \\
Recent trends annual SEER incidence rates by RUCC & 13 & 20,700 & Annual incidence rates by rural--urban category. \\
Recent trends SEER incidence trend estimates by RUCC & 17 & 3,832 & Incidence trend estimates by rural--urban category. \\
5-year age adjusted incidence rates by RUCC & 11 & 900 & Five-year age-adjusted incidence rates by rural--urban category. \\
Recent trends annual U.S. mortality rates & 11 & 218,880 & Annual U.S. mortality rates. \\
Recent trends U.S. mortality trend estimates & 15 & 30,836 & U.S. mortality trend estimates. \\
5-year age adjusted mortality rates & 10 & 9,120 & Five-year age-adjusted mortality rates. \\
Long-term trends annual U.S. mortality rates & 11 & 167,580 & Annual U.S. mortality rates. \\
Long-term trends U.S. mortality trend estimates & 15 & 14,939 & U.S. mortality trend estimates. \\
Mortality rates by age at death & 10 & 18,240 & Mortality rates stratified by age at death. \\
Median age at death & 6 & 912 & Median age at death estimates. \\
Recent trends annual U.S. mortality rates by RUCC & 13 & 12,096 & Annual U.S. mortality rates by rural--urban category. \\
Recent trends U.S. mortality trend estimates by RUCC & 17 & 1,957 & U.S. mortality trend estimates by rural--urban category. \\
5-year age adjusted mortality rates by RUCC & 11 & 612 & Five-year age-adjusted mortality rates by rural--urban category. \\
Recent trends annual SEER survival rates & 14 & 631,440 & Annual SEER survival rates. \\
Recent trends SEER survival trend estimates & 15 & 39,890 & Survival trend estimates. \\
5-year relative survival & 12 & 66,640 & Five-year relative survival estimates. \\
Long-term trends annual SEER survival rates & 13 & 48,504 & Annual SEER survival rates. \\
Long-term trends SEER survival trend estimates & 14 & 2,184 & Survival trend estimates. \\
Conditional survival & 12 & 260,160 & Conditional survival estimates. \\
Survival by survival time & 12 & 715,440 & Survival estimates by survival-time interval. \\
Complete prevalence & 7 & 752 & Prevalence. \\
Limited duration prevalence & 9 & 103,920 & Prevalence. \\
Risk of diagnosis & 9 & 179,520 & Risk-of-diagnosis estimates. \\
Risk of death & 9 & 179,520 & Risk-of-death estimates. \\
Preliminary estimates annual SEER incidence rates & 13 & 4,200 & Preliminary annual incidence rates. \\
Preliminary estimates SEER incidence trend estimates & 17 & 799 & Preliminary incidence trend estimates. \\
\end{longtable}
\endgroup

\clearpage
\section{Details on LLM-Generated Research Reports}
\label{app:llm-research-reports}

For each of the 20 cases in \benchmarkname{}, we supply the same neutral research prompt to two coding agents: Claude Code with Opus at the maximum effort setting, and Codex with GPT-5.6 Sol at the xhigh effort setting. Both agents have access to all databases and manuals included in \benchmarkname{}. The prompts specify the research topic and scope without supplying the published findings. Table~\ref{tab:report-generation-prompts} lists the exact neutral prompt used for each case.

\begingroup
\small
\setlength{\tabcolsep}{4pt}
\begin{longtable}{@{}>{\raggedright\arraybackslash}p{0.28\linewidth}>{\raggedright\arraybackslash}p{\dimexpr0.72\linewidth-2\tabcolsep\relax}@{}}
\caption{Neutral research prompts supplied to both report-generation agents for the 20 benchmark cases, in the same paper order as Table~\ref{tab:seer-twenty-reproduction}.}\label{tab:report-generation-prompts}\\
\toprule
Paper (shorter name) & Neutral research prompt \\
\midrule
\endfirsthead
\multicolumn{2}{l}{\textit{Table \thetable{} continued}}\\
\toprule
Paper (shorter name) & Neutral research prompt \\
\midrule
\endhead
\midrule
\multicolumn{2}{r}{\textit{Continued on next page}}\\
\endfoot
\bottomrule
\endlastfoot
Bone sarcoma survival by tumor site & Analyze tumor-site-specific survival trends for bone sarcomas in SEER 9 from 1980 to 2018, especially variation by histology, site, stage, age, and calendar period. \\
\addlinespace
Sex disparities in cancer incidence & Analyze sex disparities in cancer incidence across SEER 9 from 1975 to 2004, especially male-to-female incidence rate ratios by cancer site, histology, period, and age. \\
\addlinespace
Inflammatory breast cancer incidence and survival & Analyze inflammatory breast carcinoma incidence and survival trends in SEER 9 from 1988 to 2000, especially differences from locally advanced and non-T4 breast cancer by race and subtype. \\
\addlinespace
Thyroid cancer trends & Analyze thyroid cancer incidence and mortality trends in the United States from 1975 to 2009, especially by tumor size, stage, histology, age, sex, and race. \\
\addlinespace
Age at oropharyngeal cancer diagnosis & Analyze changes in age at diagnosis of oropharyngeal cancer in SEER 9 from 1975 to 2016, especially by sex, race, and human-papillomavirus-related tumor site. \\
\addlinespace
Adenoid cystic carcinoma of the breast & Analyze breast adenoid cystic carcinoma in SEER 9 from 1977 to 2006, especially incidence, clinicopathologic characteristics, subsequent primary cancers, and relative survival. \\
\addlinespace
Adult diffuse glioma incidence & Analyze incidence trends in adult diffuse gliomas in SEER 9 from 1973 to 2014, especially differences between glioblastoma and nonglioblastoma and their age, period, and birth-cohort effects. \\
\addlinespace
Childhood cancer incidence and mortality & Analyze childhood cancer incidence and mortality trends in the United States from 1975 to 1995, especially by International Classification of Childhood Cancer diagnostic group. \\
\addlinespace
Head and neck mucosal melanoma incidence & Analyze incidence and survival trends for mucosal melanoma of the head and neck in SEER 9 from 1987 to 2009, especially by anatomic site, age, sex, race, and stage. \\
\addlinespace
Esophageal adenocarcinoma incidence and mortality & Analyze temporal trends in esophageal adenocarcinoma incidence and incidence-based mortality in the United States from 1975 to 2009, especially by cancer stage, sex, and race. \\
\addlinespace
Second cancers after young-onset colorectal cancer & Analyze the risk of second primary cancers among young-onset colorectal cancer survivors in SEER 9 from 1973 to 2013, including standardized incidence ratios and absolute excess risks. \\
\addlinespace
Lung cancer incidence by sex, race, and histology & Analyze lung cancer incidence trends in SEER 9 from 1973 to 2010, especially differences by sex, race, histology, and stage distribution. \\
\addlinespace
Age disparities in colorectal cancer incidence & Analyze age-related disparities in colon and rectal cancer incidence trends in the United States from 1975 to 2010, especially differences between young adults and older patients by stage. \\
\addlinespace
Fatal prostate cancer incidence by race & Analyze trends in the incidence of fatal prostate cancer in the United States from 1975 to 2002, especially differences by race and age and changes following the introduction of PSA screening. \\
\addlinespace
Intracranial glioma incidence and mortality & Analyze intracranial glioma incidence, incidence-based mortality, and relative survival trends in SEER 9 from 1975 to 2018, especially by histology, age, sex, race, and tumor characteristics. \\
\addlinespace
Long-term mortality in young cancer survivors & Analyze 30-year mortality among five-year survivors of adolescent and young adult cancer diagnosed in SEER 9 from 1975 to 2012, distinguishing primary cancer, subsequent malignancy, and other causes of death. \\
\addlinespace
Breast cancer incidence by age and tumor features & Analyze recent trends in female breast cancer incidence in the United States, especially by age group, tumor size, stage, and estrogen/progesterone receptor status. \\
\addlinespace
Esophageal versus gastric cardia adenocarcinoma & Analyze epidemiologic differences between esophageal adenocarcinoma and gastric cardia adenocarcinoma in the nine SEER population-based registries from 1977 to 1996, especially temporal, ethnic, age, and birth-cohort patterns. \\
\addlinespace
Metastatic cancer incidence and survival & Analyze incidence trends and long-term survival of metastatic cancer in the United States from 1988 to 2018, especially trends in metastasis to specific sites such as liver, lung, bone, and brain. \\
\addlinespace
Cervical cancer incidence by race and histology & Analyze cervical adenocarcinoma and squamous cell carcinoma incidence trends among white and black women in the United States from 1976 to 2000, especially by histology, age, race, and disease stage. \\
\end{longtable}
\endgroup

\section{Perturbation Agent: Implementation Details}
\label{app:analysis-implementation}

We implement the components in Section~\ref{sec:fixed-analysis} with LLM calls and coding agents. This appendix describes their inputs, prompts, configuration, and the validation controls applied at each stage. Section~\ref{sec:results} reports the models evaluated. Our experiments include a published paper and Sol- and Opus-generated reports (Appendix~\ref{app:llm-research-reports}) for each research question.

The semantic-call wrapper can request a corrected complete JSON object after schema or parsing failures (Prompt~\ref{prompt:structured-output-repair}).

\textbf{Select Analytical Choices (Step 1).} For each published paper $p$, we identify candidate analytical choices relevant to its fully reproduced findings $F^{\mathrm{rep}}_p$. An LLM receives the paper, eligible findings, and reproduction evidence $E_p$, and extracts choices such as cohort definitions, endpoints, subgroup definitions, statistical models, and interpretation rules (Prompt~\ref{prompt:paper-frame}). For each generated report $r\in\mathcal{R}_p$, a separate call extracts choices from the report and its available execution log, linking them to the paper's eligible findings where applicable (Prompt~\ref{prompt:artifact-candidates}). We verify finding identifiers and supporting quotations, attempt to repair invalid candidates (Prompt~\ref{prompt:artifact-candidates-repair}), and remove evidence or candidates that remain ungrounded. The validated candidates from all sources are then pooled: an LLM merges descriptions of the same scientific decision and ranks the resulting choices by scientific importance, uncertainty, feasibility, and redundancy, selecting at most 20 choices before any alternative analysis is run (Prompt~\ref{prompt:canonical-panel}). Review and repair calls resolve problems with the proposed selection (Prompts~\ref{prompt:panel-review}--\ref{prompt:panel-resolution}), yielding the finalized choice set $\mathcal{C}_p$ used to propose alternatives.

\textbf{Propose Alternatives.} (Step 2 of the Perturbation Agent in Figure~\ref{fig:analysis-pipeline}) For each selected analytical choice, an LLM examines how each paper or report implements that choice, using the source text and available execution evidence (Prompt~\ref{prompt:source-mapping}) and documents the exact setting. Using these original settings and findings, another LLM proposes a defensible alternative to test across sources, adding a second only when it represents a materially different approach (Prompt~\ref{prompt:shared-alternatives}). A separate LLM reviews whether each proposed alternative is scientifically justified and identifies any conditions required for its use before execution (Prompt~\ref{prompt:alternative-review}). Choices without a defensible alternative are not tested.

Stage-specific repair calls correct invalid baseline mappings, alternatives, or alternative reviews using the associated validation errors (Prompts~\ref{prompt:source-mapping-repair}, \ref{prompt:shared-alternatives-repair}, and~\ref{prompt:alternative-review-repair}). Deterministic checks enforce valid identifiers and complete coverage of the required records.

\textbf{Execute Analyses.} (Step 3 of the Perturbation Agent in Figure~\ref{fig:analysis-pipeline}) For each choice--alternative comparison $(k,s)\in\mathcal{K}_p$, the planning LLM receives the baseline settings, the reviewed alternative, linked findings, feasibility conditions, and execution evidence; it decides whether the comparison should be executed and draws up a concrete plan for its implementation (Prompt~\ref{prompt:execution-plan}). Invalid plans are repaired against the validation errors, with deterministic checks of identifiers and comparison coverage (Prompt~\ref{prompt:execution-plan-repair}). Executable comparisons are then fed to coding agents for implementation (Prompt~\ref{prompt:analysis-execution}). The executor first runs the baseline unchanged and checks it against the frozen validation criteria. It then copies the baseline, changes only the specified choice and its listed dependent calculations, and saves the baseline output, alternative output, validation evidence, and exact code diff. These saved artifacts form the comparison evidence $e_k(s)$ returned by the Execute Analyses stage in Algorithm~\ref{alg:fixed-analysis}.

The pipeline prepares the \benchmarkname{} environment for execution. The launch prompt (Prompt~\ref{prompt:executor-launch}) directs the coding agent to follow the execution instructions (Prompt~\ref{prompt:analysis-execution}). The agent recovers the baseline implementation from preserved code, structured execution traces, or legacy transcripts, in that order of preference; it constructs a fresh implementation only when no prior implementation evidence survives. It then validates the unchanged baseline, executes the specified alternatives, and saves the results and supporting evidence. If a Codex session ends successfully without the required results file, a continuation prompt asks it to finish using the same workspace (Prompt~\ref{prompt:executor-continuation}). If the file remains missing, a final prompt stops further analysis and requires it to record existing results, marking unfinished comparisons as unavailable (Prompt~\ref{prompt:executor-finalization}).

\textbf{Assess Consequences.} (Step 4 of the Perturbation Agent in Figure~\ref{fig:analysis-pipeline}) A separate assessment LLM receives each comparison, its eligibility evidence, frozen comparison criteria, and the saved execution evidence $e_k(s)$ (Prompt~\ref{prompt:adjudication}). This implements the Assess Consequences stage in Algorithm~\ref{alg:fixed-analysis}, assigning an outcome $y_k(s)$ using the \texttt{AnalysisOutcome} type defined in Equation~\ref{eq:analysis-decision}.

For \texttt{DirectChange} outcomes, a further LLM call assigns a scientific-consequence score from 1 to 9 based on the report's conclusions and the baseline-to-alternative evidence (Prompt~\ref{prompt:consequentiality-ordinal-v3}), adapted from GRADE's nine-point outcome-importance framework \citep{schunemann2008grade}. The score bands are interpreted as described in Section~\ref{sec:fixed-analysis}.

\section{Reproduction Outcome Breakdowns in \benchmarkname{}}
\label{app:reproduction}

Table~\ref{tab:seer-twenty-reproduction-counts} provides the counts and percentages for all three reproduction outcomes for each paper, alongside its numbers of findings and atomic claims.

\begin{table}[p]
\centering
\caption{Tabular breakdown of the reproduction outcomes shown in Table~\ref{tab:seer-twenty-reproduction}, in the same paper order. Each outcome cell gives the number of atomic claims and the percentage of that paper\textquotesingle s  atomic claims. Findings counts refer to distinct parent findings represented by these claims.}
\label{tab:seer-twenty-reproduction-counts}
\footnotesize
\setlength{\tabcolsep}{3pt}
\renewcommand{\arraystretch}{1.18}
\begin{tabularx}{\linewidth}{@{}>{\raggedright\arraybackslash}Xrrrrr@{}}
\toprule
Paper (shorter name) & Findings & \shortstack{Atomic\\claims} & Reproduced & Close & \shortstack{Not\\reproduced} \\
\midrule
Bone sarcoma survival by tumor site & 56 & 188 & 187 (99.5\%) & 0 (0.0\%) & 1 (0.5\%) \\
Sex disparities in cancer incidence & 36 & 151 & 140 (92.7\%) & 1 (0.7\%) & 10 (6.6\%) \\
Inflammatory breast cancer incidence and survival & 40 & 131 & 119 (90.8\%) & 2 (1.5\%) & 10 (7.6\%) \\
Thyroid cancer trends & 33 & 91 & 80 (87.9\%) & 1 (1.1\%) & 10 (11.0\%) \\
Age at oropharyngeal cancer diagnosis & 75 & 154 & 129 (83.8\%) & 9 (5.8\%) & 16 (10.4\%) \\
Adenoid cystic carcinoma of the breast & 51 & 128 & 103 (80.5\%) & 2 (1.6\%) & 23 (18.0\%) \\
Adult diffuse glioma incidence & 60 & 111 & 89 (80.2\%) & 4 (3.6\%) & 18 (16.2\%) \\
Childhood cancer incidence and mortality & 52 & 144 & 113 (78.5\%) & 6 (4.2\%) & 25 (17.4\%) \\
Head and neck mucosal melanoma incidence & 20 & 54 & 42 (77.8\%) & 1 (1.9\%) & 11 (20.4\%) \\
Esophageal adenocarcinoma incidence and mortality & 33 & 52 & 40 (76.9\%) & 3 (5.8\%) & 9 (17.3\%) \\
Second cancers after young-onset colorectal cancer & 59 & 124 & 95 (76.6\%) & 14 (11.3\%) & 15 (12.1\%) \\
Lung cancer incidence by sex, race, and histology & 36 & 111 & 80 (72.1\%) & 7 (6.3\%) & 24 (21.6\%) \\
Age disparities in colorectal cancer incidence & 27 & 82 & 59 (72.0\%) & 7 (8.5\%) & 16 (19.5\%) \\
Fatal prostate cancer incidence by race & 32 & 63 & 45 (71.4\%) & 2 (3.2\%) & 16 (25.4\%) \\
Intracranial glioma incidence and mortality & 67 & 94 & 66 (70.2\%) & 2 (2.1\%) & 26 (27.7\%) \\
Long-term mortality in young cancer survivors & 62 & 178 & 123 (69.1\%) & 5 (2.8\%) & 50 (28.1\%) \\
Breast cancer incidence by age and tumor features & 31 & 67 & 39 (58.2\%) & 3 (4.5\%) & 25 (37.3\%) \\
Esophageal versus gastric cardia adenocarcinoma & 34 & 51 & 29 (56.9\%) & 2 (3.9\%) & 20 (39.2\%) \\
Metastatic cancer incidence and survival & 51 & 96 & 52 (54.2\%) & 4 (4.2\%) & 40 (41.7\%) \\
Cervical cancer incidence by race and histology & 46 & 120 & 64 (53.3\%) & 1 (0.8\%) & 55 (45.8\%) \\
\midrule
\textbf{Total} & \textbf{901} & \textbf{2,190} & \textbf{1,694 (77.4\%)} & \textbf{76 (3.5\%)} & \textbf{420 (19.2\%)} \\
\bottomrule
\end{tabularx}
\end{table}

Table~\ref{tab:seer-twenty-reproduction-breakdown} reports the 20-paper breakdown using cumulative 1\%, 5\%, and 10\% tolerances for quantitative claims and binary outcomes for qualitative claims. Of 1,695 quantitative claims, 1,299 are reproduced (76.6\%), 76 are close, and 320 are not reproduced. Of 495 qualitative claims, 395 are reproduced (79.8\%) and 100 are not reproduced. Together these yield the 1,694 reproduced, 76 close, and 420 not-reproduced claims reported in the main text.

\begin{table}[p]
\centering
\caption{Atomic reproduction outcomes by claim type for the 20-paper \benchmarkname{} benchmark.}
\label{tab:seer-twenty-reproduction-breakdown}
\fontsize{5.0}{5.8}\selectfont
\setlength{\tabcolsep}{1.5pt}
\renewcommand{\arraystretch}{1.0}
\begin{tabularx}{\linewidth}{@{}>{\raggedright\arraybackslash}p{0.245\linewidth}r*{6}{>{\raggedleft\arraybackslash}X}@{}}
\toprule
 & & \multicolumn{4}{c}{Quantitative claims} & \multicolumn{2}{c}{Qualitative claims} \\
\cmidrule(lr){3-6}\cmidrule(lr){7-8}
Paper & Cit. & \shortstack{1\%\\tolerance} & \shortstack{5\%\\tolerance} & \shortstack{10\%\\tolerance} & All & Reprod. & Not reprod. \\
\midrule
Bone sarcoma survival by tumor site & 37 & 119 (88.1\%) & 135 (100.0\%) & 135 (100.0\%) & 135 (100.0\%) & 52 (98.1\%) & 1 (1.9\%) \\
Sex disparities in cancer incidence & 466 & 87 (75.0\%) & 114 (98.3\%) & 115 (99.1\%) & 116 (100.0\%) & 26 (74.3\%) & 9 (25.7\%) \\
Inflammatory breast cancer incidence and survival & 702 & 78 (74.3\%) & 98 (93.3\%) & 100 (95.2\%) & 105 (100.0\%) & 21 (80.8\%) & 5 (19.2\%) \\
Thyroid cancer trends & 1,628 & 52 (61.9\%) & 74 (88.1\%) & 75 (89.3\%) & 84 (100.0\%) & 6 (85.7\%) & 1 (14.3\%) \\
Age at oropharyngeal cancer diagnosis & 29 & 101 (73.2\%) & 117 (84.8\%) & 126 (91.3\%) & 138 (100.0\%) & 12 (75.0\%) & 4 (25.0\%) \\
Adenoid cystic carcinoma of the breast & 188 & 75 (64.7\%) & 92 (79.3\%) & 94 (81.0\%) & 116 (100.0\%) & 11 (91.7\%) & 1 (8.3\%) \\
Adult diffuse glioma incidence & 72 & 56 (65.9\%) & 67 (78.8\%) & 71 (83.5\%) & 85 (100.0\%) & 22 (84.6\%) & 4 (15.4\%) \\
Childhood cancer incidence and mortality & 438 & 56 (52.3\%) & 84 (78.5\%) & 90 (84.1\%) & 107 (100.0\%) & 29 (78.4\%) & 8 (21.6\%) \\
Head and neck mucosal melanoma incidence & 83 & 28 (56.0\%) & 38 (76.0\%) & 39 (78.0\%) & 50 (100.0\%) & 4 (100.0\%) & 0 (0.0\%) \\
Esophageal adenocarcinoma incidence and mortality & 510 & 15 (40.5\%) & 25 (67.6\%) & 28 (75.7\%) & 37 (100.0\%) & 15 (100.0\%) & 0 (0.0\%) \\
Second cancers after young-onset colorectal cancer & 43 & 31 (32.3\%) & 70 (72.9\%) & 84 (87.5\%) & 96 (100.0\%) & 25 (89.3\%) & 3 (10.7\%) \\
Lung cancer incidence by sex, race, and histology & 368 & 51 (52.0\%) & 69 (70.4\%) & 76 (77.6\%) & 98 (100.0\%) & 11 (84.6\%) & 2 (15.4\%) \\
Age disparities in colorectal cancer incidence & 1,102 & 20 (39.2\%) & 29 (56.9\%) & 36 (70.6\%) & 51 (100.0\%) & 30 (96.8\%) & 1 (3.2\%) \\
Fatal prostate cancer incidence by race & 111 & 28 (60.9\%) & 31 (67.4\%) & 33 (71.7\%) & 46 (100.0\%) & 14 (82.4\%) & 3 (17.6\%) \\
Intracranial glioma incidence and mortality & 116 & 53 (58.2\%) & 63 (69.2\%) & 65 (71.4\%) & 91 (100.0\%) & 3 (100.0\%) & 0 (0.0\%) \\
Long-term mortality in young cancer survivors & 23 & 65 (47.1\%) & 98 (71.0\%) & 103 (74.6\%) & 138 (100.0\%) & 25 (62.5\%) & 15 (37.5\%) \\
Breast cancer incidence by age and tumor features & 352 & 19 (37.3\%) & 26 (51.0\%) & 29 (56.9\%) & 51 (100.0\%) & 13 (81.2\%) & 3 (18.8\%) \\
Esophageal versus gastric cardia adenocarcinoma & 266 & 7 (25.0\%) & 10 (35.7\%) & 12 (42.9\%) & 28 (100.0\%) & 19 (82.6\%) & 4 (17.4\%) \\
Metastatic cancer incidence and survival & 68 & 10 (15.6\%) & 26 (40.6\%) & 30 (46.9\%) & 64 (100.0\%) & 26 (81.2\%) & 6 (18.8\%) \\
Cervical cancer incidence by race and histology & 448 & 25 (42.4\%) & 33 (55.9\%) & 34 (57.6\%) & 59 (100.0\%) & 31 (50.8\%) & 30 (49.2\%) \\
\midrule
\textbf{Total} & & \textbf{976 (57.6\%)} & \textbf{1,299 (76.6\%)} & \textbf{1,375 (81.1\%)} & \textbf{1,695 (100.0\%)} & \textbf{395 (79.8\%)} & \textbf{100 (20.2\%)} \\
\bottomrule
\end{tabularx}

\end{table}

\section{Reproduction Grading Details}
\label{app:benchmark-construction}

Findings are extracted (Prompt~\ref{prompt:finding-extraction}), decomposed into atomic targets (Prompt~\ref{prompt:atomic-decomposition}), and attempted to be reproduced (Prompt~\ref{prompt:reproduction-execution}). The reproduction grader separates evidence assessment from numerical comparison. An LLM extracts the result reported for each atomic claim and assesses its method and answer provenance (Prompt~\ref{prompt:atomic-grading}); deterministic code then applies the numerical grading policy. The reproduction agent's own statement that a finding was reproduced does not determine the final verdict.

\subsection{Judge Inputs and Evidence Assessment}
\label{app:reproduction-evidence}

\textbf{Inputs and unit of assessment.} One judge call receives the reproduction report for an original finding together with its atomic targets. Each target includes its identifier, claim text, reference answer, numerical or semantic evaluation mode, verification target, source quotation, and required analytical assumptions. The judge is instructed to assess every atomic target independently: reproducing one component of a finding does not establish that its other components were reproduced. The grading call receives the report and structured targets, rather than independently inspecting the complete code and execution trace or rerunning the analysis.

\textbf{Result availability.} For each target, the judge records whether the requested result is reported, not reported, or unclear. A missing reproduction report is assigned a not-reproduced verdict directly. When a report exists but omits the requested result or provides ambiguous evidence, the target likewise receives no reproduction credit.

\textbf{Method validity.} The judge assesses whether the reported computation uses a valid method for the specific target. This includes compatibility with required cohort bounds, disease and variable coding, outcome measures, grouping, denominators, standardization conventions, temporal data definitions, and statistical-model settings. A conflicting analytical assumption makes the method invalid; insufficient methodological evidence makes it unclear. The report need not repeat every required assumption verbatim, but it must establish a compatible computation. Both invalid and unclear methods lead to a not-reproduced verdict, even when the extracted numerical value matches the reference.

\textbf{Answer provenance.} The judge also flags evidence that the result was obtained by recalling, copying, or searching for the published answer instead of computing it from the supplied data. Such results receive no reproduction credit. Consulting documentation, variable definitions, or coding guidance is permitted when the agent subsequently performs the computation. An exact numerical match alone is not evidence of copying.

\subsection{Numerical Extraction and Comparison}

Only claims passing the evidence checks in Section~\ref{app:reproduction-evidence} proceed to numerical grading. The process has three steps.

\textbf{1. Extract the computed result.} The LLM extracts each required value, its unit, and its relation ($=$, $<$, $\leq$, $>$, or $\geq$), using unrounded values when available. It follows the target's requested representation---for example, distinguishing the magnitude of a decline from a signed change. It does not calculate numerical error or assign the numerical verdict.

\textbf{2. Compare each required component.} For an ordinary point estimate with reference value $t\ne0$ and computed value $x$, deterministic code calculates
\[
\varepsilon=\frac{|x-t|}{|t|}.
\]
The component is \textbf{Reproduced} if $\varepsilon\leq5\%$, \textbf{Close} if $5\%<\varepsilon\leq10\%$, and \textbf{Not reproduced} otherwise. Counts also use this rule unless the target specifies another tolerance.

The following cases modify this default comparison:

\begingroup
\renewcommand{\arraystretch}{1.15}
\noindent\begin{tabularx}{\linewidth}{@{}>{\raggedright\arraybackslash}p{0.21\linewidth}>{\raggedright\arraybackslash}X@{}}
\toprule
\textbf{Case} & \textbf{Comparison rule} \\
\midrule
\textbf{Rounding allowance} & If the target specifies its reported decimal precision, accept differences up to half a unit in the last reported decimal place---for example, $0.05$ for one decimal place. \\
\addlinespace
\textbf{Explicit tolerance} & Use the target's exact-equality or absolute-error thresholds instead of the default relative thresholds, subject to the rounding allowance. \\
\addlinespace
\textbf{Zero reference} & Under the default relative rule, require $x=0$, unless the rounding allowance applies. \\
\addlinespace
\textbf{Inequality} & Check whether the computed value satisfies the target inequality. If only a bound is reported, it must imply the target bound. A bound cannot establish a required point value. These comparisons have no Close category. \\
\addlinespace
\textbf{Interval} & For an acceptable range, check whether the computed scalar lies within its boundaries, respecting strict or inclusive limits. For a reported interval, compare its two endpoints separately. \\
\bottomrule
\end{tabularx}
\endgroup

\textbf{3. Combine the component verdicts.} Required components are matched in their declared order; optional uncertainty components are excluded. The least favorable component verdict determines the atomic claim's verdict. Missing or unparseable required values yield ``Not reproduced''.

\subsection{Grading Qualitative Claims}

For qualitative or categorical claims, the judge determines whether the reproduced result supports the same conclusion as the published claim. A claim is marked reproduced only when the conclusions are equivalent and the evidence passes the availability, method-validity, and answer-provenance checks described above. Approximately equivalent, conflicting, or unclear conclusions are marked not reproduced. Unlike numerical claims, qualitative claims have no close category.

\paragraph{Implementation checks and saved records.}
The implementation verifies that the judge returns the requested claim identifiers and that extracted numerical results have the required structure. Saved records include the extracted result, assessment fields, rationale, final verdict, and grading-policy version. Numerical thresholds can therefore be reapplied to saved extractions without another judge call. Table~\ref{tab:seer-twenty-reproduction-breakdown} reports outcomes by claim type.

\paragraph{Scope of the grades.}
These grades assess whether the reproduction report provides sufficient evidence for the original claim. They do not independently verify the complete analysis code or establish that undocumented analytical choices match those of the original authors.

\clearpage
\section{Full Titles of Benchmark Papers}
\label{app:paper-titles}

Table~\ref{tab:paper-titles} maps the shorter names used in Table~\ref{tab:seer-twenty-reproduction} to the full titles of all 20 papers in \benchmarkname{}, in the same order.

\begingroup
\small
\setlength{\tabcolsep}{4pt}
\renewcommand{\arraystretch}{1.1}
\begin{longtable}{@{}p{0.28\linewidth}p{\dimexpr0.72\linewidth-2\tabcolsep\relax}@{}}
\caption{Full titles corresponding to the shorter paper names in Table~\ref{tab:seer-twenty-reproduction}.}\label{tab:paper-titles}\\
\toprule
Shorter name & Full paper title \\
\midrule
\endfirsthead
\toprule
Shorter name & Full paper title \\
\midrule
\endhead
\bottomrule
\endfoot
Bone sarcoma survival by tumor site & Trends in Tumor Site-Specific Survival of Bone Sarcomas from 1980 to 2018: A Surveillance, Epidemiology and End Results-Based Study \\
\addlinespace[3pt]
Sex disparities in cancer incidence & Sex Disparities in Cancer Incidence by Period and Age \\
\addlinespace[3pt]
Inflammatory breast cancer incidence and survival & Trends in Inflammatory Breast Carcinoma Incidence and Survival: The Surveillance, Epidemiology, and End Results Program at the National Cancer Institute \\
\addlinespace[3pt]
Thyroid cancer trends & Current Thyroid Cancer Trends in the United States \\
\addlinespace[3pt]
Age at oropharyngeal cancer diagnosis & Change in Age at Diagnosis of Oropharyngeal Cancer in the United States, 1975--2016 \\
\addlinespace[3pt]
Adenoid cystic carcinoma of the breast & Adenoid cystic carcinoma of the breast in the United States (1977 to 2006): a population-based cohort study \\
\addlinespace[3pt]
Adult diffuse glioma incidence & Trends and patterns of incidence of diffuse glioma in adults in the United States, 1973-2014 \\
\addlinespace[3pt]
Childhood cancer incidence and mortality & Cancer Surveillance Series: Recent Trends in Childhood Cancer Incidence and Mortality in the United States \\
\addlinespace[3pt]
Head and neck mucosal melanoma incidence & Rising Incidence of Mucosal Melanoma of the Head and Neck in the United States \\
\addlinespace[3pt]
Esophageal adenocarcinoma incidence and mortality & Trends in Esophageal Adenocarcinoma Incidence and Mortality \\
\addlinespace[3pt]
Second cancers after young-onset colorectal cancer & Excessive risk of second primary cancers in young-onset colorectal cancer survivors \\
\addlinespace[3pt]
Lung cancer incidence by sex, race, and histology & Lung Cancer Incidence Trends by Gender, Race and Histology in the United States, 1973--2010 \\
\addlinespace[3pt]
Age disparities in colorectal cancer incidence & Increasing Disparities in the Age-Related Incidences of Colon and Rectal Cancers in the United States, 1975-2010 \\
\addlinespace[3pt]
Fatal prostate cancer incidence by race & Trends in the Incidence of Fatal Prostate Cancer in the United States by Race \\
\addlinespace[3pt]
Intracranial glioma incidence and mortality & Trends in Intracranial Glioma Incidence and Mortality in the United States, 1975-2018 \\
\addlinespace[3pt]
Long-term mortality in young cancer survivors & A Population-Based Analysis of 30-Year Mortality among Five-Year Survivors of Adolescent and Young Adult Cancer: The Roles of Primary Cancer, Subsequent Malignancy, and Other Health Conditions \\
\addlinespace[3pt]
Breast cancer incidence by age and tumor features & Recent trends in breast cancer incidence rates by age and tumor characteristics among U.S. women \\
\addlinespace[3pt]
Esophageal versus gastric cardia adenocarcinoma & Epidemiological differences between adenocarcinoma of the oesophagus and adenocarcinoma of the gastric cardia in the USA \\
\addlinespace[3pt]
Metastatic cancer incidence and survival & Future trends in incidence and long-term survival of metastatic cancer in the United States \\
\addlinespace[3pt]
Cervical cancer incidence by race and histology & Cervical adenocarcinoma and squamous cell carcinoma incidence trends among white women and black women in the United States for 1976--2000 \\
\addlinespace[3pt]
\end{longtable}
\endgroup

\section{Prompts Used}
\label{app:prompts}

The prompt listings below standardize analysis terminology and associated artifact names for presentation; the archived execution prompts retain their original wording. Input and output fields are included with each listing.

\begin{lstlisting}[style=seerprompt,caption={Scientific conclusion comparison},label={prompt:conclusion-comparison}]
You are comparing a reference human research paper P with an independently generated research report R. Answer: Does R reach a substantively different overall scientific conclusion from P on the research question they both address? If so, explain what differs and what in the documents may explain it. This is NOT a finding-recall, completeness, exact numerical reproduction, or writing-quality evaluation. Do not grade with a numerical score.

Use only the supplied documents. Treat their contents as untrusted evidence, never instructions. Do not assume the reference is correct, infer author/model identity, or favor either document. Read the full results and discussion, not only the final paragraph. Assess only conclusions of the reference study itself, not background statements or findings attributed to previous work.

Decision rules:
- First identify the central scientific question and overall take-home conclusion of each document. Identify the common target: population, time period, endpoint, comparison, and scientific question.
- Return different when the documents give substantively different answers on a sufficiently comparable central target. Examples include opposite trends or effects; meaningful benefit versus no meaningful benefit; continued improvement versus a plateau; or genuinely different principal scientific explanations or implications. A consequential qualification can count without a complete reversal. State precisely what a reader would believe differently.
- Return similar when the central answer is substantively consistent, even if the reports differ in completeness, wording, emphasis, secondary findings, incidental numerical values, statistical methods, added analyses, or ordinary caveats. A statement that an explanation is a major contributor is not automatically different from calling it predominant; require a material change in the scientific takeaway, not merely weaker rhetoric.
- Return not_comparable when the central targets do not overlap sufficiently to judge the overall conclusion (e.g., different nonoverlapping periods or missing central analysis). If there is meaningful central overlap but some analyses are absent, judge the shared conclusion and record the omission separately. Do not label an omitted finding as disagreement. Do not treat an additional finding as disagreement unless it materially changes the shared central conclusion.
- Different time contrasts or subgroup definitions can reconcile apparently opposite statements: stability over a full period and decline over its later segment can both hold. Do not call these contradictory without checking their targets. Statistical nonsignificance alone is not evidence of the opposite effect; different p-values alone need not change the conclusion.
- Keep empirical differences (what happened) separate from interpretive differences (why it happened or what it implies). Report the type of substantive difference only if the overall verdict is different.
- Explain WHY cautiously. Text can establish that the analyses used different cohorts, endpoints, time windows, definitions, models, adjustments, or interpretive assumptions; it usually cannot establish that such a choice caused a conclusion difference. Separate a documented analytical difference from a plausible causal explanation. Do not speculate about coding errors, model bias, missing variables, or mechanisms absent from the documents. Use insufficient_information when the reason cannot be established. This exercise does not validate execution or establish which conclusion is true.

Return only JSON with this exact structure:
{
 "verdict": "similar" | "different" | "not_comparable",
 "human_takeaway": "Brief central conclusion of the current reference study.",
 "report_takeaway": "Brief central conclusion of the report.",
 "shared_question_and_scope": "What can actually be compared, including material scope differences.",
 "difference_type": "none" | "empirical" | "interpretive" | "both" | "not_comparable",
 "substantive_differences": [
  {
   "difference": "A material difference in the shared central conclusion, not merely an omitted or secondary finding.",
   "type": "empirical" | "interpretive",
   "human_quote": "Short contiguous verbatim reference passage.",
   "report_quote": "Short contiguous verbatim report passage.",
   "why_material": "What a reader would conclude differently and why this is not just scope, emphasis, or omission."
  }
 ],
 "explanation": [
  {
   "factor": "study_period" | "population_or_cohort" | "endpoint_or_definition" | "statistical_method_or_adjustment" | "interpretation_or_causal_attribution" | "other_documented_factor" | "insufficient_information",
   "observation": "What the documents establish about a possible explanation for the different conclusions or inability to compare.",
   "evidential_status": "documented_difference_causal_role_unproven" | "plausible_explanation" | "insufficient_information",
   "human_quote": "Short contiguous verbatim passage, or null if absent.",
   "report_quote": "Short contiguous verbatim passage, or null if absent.",
   "causal_limit": "What cannot be established about why the conclusions differ from these texts alone."
  }
 ],
 "coverage_or_emphasis_notes": "Omissions, scope shifts, secondary differences, or caveats that should not themselves be counted as substantive disagreement.",
 "overall_evidence": {
  "human_quote": "Short contiguous verbatim passage supporting the human takeaway.",
  "report_quote": "Short contiguous verbatim passage supporting the report takeaway."
 },
 "rationale": "Two to four sentences justifying the overall decision."
}

For similar: difference_type=none, substantive_differences=[], explanation=[] (put non-substantive distinctions in coverage_or_emphasis_notes).
For different: difference_type=empirical, interpretive, or both consistently with listed differences; provide 1–3 substantive_differences and 1–3 explanation entries. If reasons are unknown, use insufficient_information rather than invent them.
For not_comparable: difference_type=not_comparable, substantive_differences=[], and 1–3 explanation entries about scope or missing central analysis.
Every quotation must be an exact contiguous source substring after whitespace normalization, preferably 10–35 words and at most 70 words. Retain Markdown formatting within quoted text. Do not insert ellipses or join separated passages. Only explanation quotations may be null. For documented differences or plausible explanations, provide evidence from both documents; for insufficient_information, null is allowed. Do not fill fields with unsupported causal claims.

Input fields:
  reference_paper: full paper text
  analysis_report: full report text

Output fields:
{
  verdict: "similar" | "different" | "not_comparable",
  human_takeaway: string,
  report_takeaway: string,
  shared_question_and_scope: string,
  difference_type: "none" | "empirical" | "interpretive" | "both" | "not_comparable",
  substantive_differences: [{
    difference: string,
    type: "empirical" | "interpretive",
    human_quote: string,
    report_quote: string,
    why_material: string
  }],
  explanation: [{
    factor: "study_period" | "population_or_cohort" | "endpoint_or_definition" | "statistical_method_or_adjustment" | "interpretation_or_causal_attribution" | "other_documented_factor" | "insufficient_information",
    observation: string,
    evidential_status: "documented_difference_causal_role_unproven" | "plausible_explanation" | "insufficient_information",
    human_quote: string | null,
    report_quote: string | null,
    causal_limit: string
  }],
  coverage_or_emphasis_notes: string,
  overall_evidence: {human_quote: string, report_quote: string},
  rationale: string
}
\end{lstlisting}
\raggedbottom

\begin{lstlisting}[style=seerprompt,caption={Finding extraction},label={prompt:finding-extraction}]
You are a meticulous oncology-epidemiology research assistant. You are given the full text of a published study that analyzed the SEER cancer registry. Your job is to extract a bank of ATOMIC VERIFIABLE CLAIMS: the study's concrete, data-derived findings and reproduction-relevant study context, each stated so it could be independently checked from the raw data or used to reproduce a data-derived result.

Rules for an atomic verifiable claim:
- It states ONE verifiable claim only. Split compound findings into separate claims.
- It may be numeric or non-numeric, but it must be empirically checkable from the study data or necessary for reproducing such a check.
- If the paper reports concrete numbers for the claim, include them in "numbers" with units and context.
- If the claim is qualitative but data-derived, such as a distribution shape, ordering, rank, threshold crossing, curve pattern, subgroup comparison, table pattern, or figure pattern, keep "numbers" as an empty list and describe what should be checked in "verification_target".
- Also extract operational study context when it is necessary to reproduce findings: cohort definitions, SEER variable/code definitions, inclusion/exclusion criteria, grouping rules, endpoint definitions, statistical methods, plotting methods, adjustment methods, and table/figure construction rules.
- Preserve enough detail that another agent could decide what data subset, variables, grouping, transformation, model, plot, or comparison is needed.
- Capture the verbatim sentence(s), table row, table note, figure caption, or methods text that support the claim.
- Extract findings and reproducibility context from results, tables, figures, figure captions, table notes, and methods sections.
- Skip background, motivation, citations of other studies, unsupported interpretation/speculation, and purely clinical discussion not tied to a reproducible data operation.

Return a JSON object with a single key "claims" whose value is a LIST of objects. Each object MUST use exactly these field names, and no others:
- "claim": string -- the one-sentence claim, written to be checkable.
- "evidence_type": one of "numeric", "categorical", "distributional", "trend_shape", "ranking", "table_fact", "figure_pattern", "method_context", "other".
- "numbers": a LIST of objects, each with "value" (string), "unit" (string), and "context" (string). Use an empty list if the claim is non-numeric.
- "verification_target": string -- what an analysis agent should compute, plot, compare, inspect, or apply to verify this claim or use it during reproduction.
- "source_quote": string -- the verbatim supporting sentence(s), table row, table note, figure caption, or methods text from the paper.

Use these exact keys. Do not rename fields. Return only the JSON object, no prose or markdown fences.

Paper title: {{title}}

Full text:
{{full_text}}


Input fields:
  title: paper title
  full_text: full paper text

Output fields:
{
  claims: [{
    claim: string,
    evidence_type: "numeric" | "categorical" | "distributional" | "trend_shape" | "ranking" | "table_fact" | "figure_pattern" | "method_context" | "other",
    numbers: [{value: string, unit: string, context: string}],
    verification_target: string,
    source_quote: string
  }]
}
\end{lstlisting}

\begin{lstlisting}[style=seerprompt,caption={Atomic claim decomposition},label={prompt:atomic-decomposition}]
You are splitting ONE grouped SEER reproduction claim into independently
testable scientific findings.

For numeric results, "atomic" means one substantive reported scalar. Related
numbers retain shared context but become separate atomic claims so each can be
reproduced and graded independently.

Splitting rules:
- Preserve every independently testable scientific result in the source claim.
- The curated source_claim.numbers array is the complete numeric target
  inventory. Every numeric child must correspond to an entry in that array.
  Never create a target from a number that appears only in source_quote or the
  full paper text.
- Split every substantive point value into its own atomic claim, including each
  baseline/follow-up value, each subgroup value, numerators, denominators,
  derived percentages, absolute differences, ratios, and other reported results.
- Example: "1,524 of 3,223 cases, or 47%" becomes three atomic claims: 1,524,
  3,223, and 47%. None is merely optional support.
- Example: an increase from 4.9 to 14.3 becomes one claim for 4.9 and one claim
  for 14.3. Preserve the year and baseline/follow-up role in each claim.
- Keep a confidence interval with its point estimate in one atomic claim. Never
  turn confidence-interval endpoints into standalone claims.
- Treat any parenthetical interval immediately following and enclosing a point
  estimate as optional uncertainty, even when source_claim.numbers calls it a
  "range" rather than a confidence interval. Example: `9.4 (8.9-9.9)` is one
  atomic claim for 9.4 with optional lower=8.9 and upper=9.9, not a second
  required range claim.
- Keep a meaningful range together as one atomic claim. Never turn range
  endpoints into standalone claims.
- Treat cohort definitions and analysis settings as context, not independent
  findings. Study years, age eligibility, registry counts, stage codes, and
  similar design numbers belong in required_analysis_assumptions unless the
  grouped claim explicitly presents that item as the substantive result.
- Split genuinely independent outcomes, subgroups, tumor types, or conclusions
  when each can stand alone scientifically.
- Create a qualitative direction or comparison claim only when that conclusion
  is independently testable and not redundant with the numeric child claims.
- Do not invent facts, precision, cohorts, units, or conclusions.
- Keep enough cohort, outcome, subgroup, geography, and time context for each atomic claim to stand alone.
- Use the full paper to resolve context needed to make the exact result
  independently reproducible, but include only cohort and analysis details that
  apply to that result; do not automatically propagate broad paper-level
  restrictions into every atomic claim.
- A numeric atomic answer normally contains exactly one required component and
  exactly one substantive number in answer.value. Do not include contextual
  design numbers or other substantive results in answer.value.
- A genuinely range-valued result is the sole exception: keep its lower and
  upper endpoints together as two required components with roles "range_lower"
  and "range_upper". Set range_semantics="reported_interval" when the paper
  reports both endpoints as the result and both must be reproduced. Set
  range_semantics="acceptable_interval" when the range is a criterion and any
  scalar inside it satisfies the result. Do not use this exception for
  before/after or group pairs. Otherwise set range_semantics=null.
- Example: "the IRR declined to 4-5 at the oldest ages" is an
  acceptable_interval. A computed oldest-age IRR of 4.814 satisfies it; the
  agent need not manufacture or reproduce endpoints 4 and 5.
- Every required substantive component uses role "primary" except the two
  endpoints of a genuinely range-valued result.
- Preserve the numeric relation for every component. Use relation "eq" for a
  point estimate or an endpoint that the agent must reproduce, "lt" for
  "less than" or "below", "lte" for "at most" or "not exceeding", "gt" for
  "more than" or "above", and "gte" for "at least" or "no less than".
  Never discard an inequality or turn a lower/upper bound into an equality.
- Relation and comparator are separate: relation describes which values satisfy
  the finding, while comparator describes exact versus tolerance-based numeric
  matching for equality targets.
- Use comparator "relative_numeric" for approximate estimates, percentages,
  rates, ratios, and confidence limits. Use "exact_numeric" for exact counts or
  genuinely exact result parameters.
- Every required numeric component must include tolerance. Use
  {"kind":"relative","reproduced":0.05,"close":0.10} for estimates, rates,
  percentages, ratios, APCs, p-values, and counts. Use an absolute tolerance
  appropriate to the paper's precision for years, ages, timing, and targets at
  or near zero; for example, an approximate turning year reported to the nearest
  year normally uses {"kind":"absolute","reproduced":1,"close":2}. Use
  {"kind":"exact","reproduced":0,"close":0} only when exact identity is
  scientifically required.
- Set reported_decimal_places to the number of decimal places visibly reported
  for an equality target, so an unrounded reproduction that rounds to the paper's
  displayed value can match. Set it to null for inequalities and values whose
  displayed precision is not meaningful. Do not use this field to make vague
  words such as "about" artificially permissive.
- Use one canonical representation for changes and keep claim, label, value,
  and component consistent. "Decreased by 5%" and "magnitude of the decline"
  require the positive magnitude 5. "Increased by 5%" requires positive 5.
  "APC was -5%" or "signed APC" requires signed -5. Never pair change-magnitude
  wording with a negative target. Preserve threshold relations on the magnitude:
  "the magnitude of the decline was less than 1%" stores positive 1 with
  relation="lt".
- Keep confidence or uncertainty bounds with their point estimate in
  answer.uncertainty, not as separate atomic claims and not in components.
  uncertainty contains kind, level, lower, and upper. Use kind
  "interquartile_range" for an IQR. Each bound uses the same fields as a
  component but must have required=false. A correct point estimate can therefore
  reproduce the finding even when its uncertainty is not reported.
- Use answer_type "continuous_numeric" when all required result values are
  percentages, rates, APCs, p-values, ratios, estimates, confidence limits, or
  other approximate numeric values.
- Use answer_type "exact_numeric" only when the substantive result is a fixed
  numeric parameter or meaningful exact range.
- Use answer_type "count" when all required result values are exact counts.
- Use answer_type "categorical" for exact named categories or labels.
- Use answer_type "qualitative" for directions, rankings, comparisons, and patterns requiring semantic grading.
- Copy or closely preserve the source quote that supports each atomic claim.
- Populate required_analysis_assumptions with the answer-safe protocol needed
  to compute this exact result. This setup is shown to the QA agent separately
  from the question, so it may be more specific than the source sentence.
- In required_analysis_assumptions.cohort, preserve every applicable registry,
  diagnosis-year, sex, race, stage, histology, behavior, primary-sequence,
  reporting-source, and eligibility restriction. Always state applicable age
  bounds explicitly and preserve inclusive semantics, for example "Include
  women age >=40 years" or "Restrict to ages 45-84 years, inclusive".
- In required_analysis_assumptions.coding, record the paper's exact variable,
  recode, code system and edition, code list, precedence rule, missing-value
  handling, and exclusion logic when stated.
- In required_analysis_assumptions.denominator_and_standardization, record the
  denominator population, rate scale, standard population, age bins, weight
  handling, and whether weights are renormalized after an age restriction.
- In required_analysis_assumptions.temporal_data, record the data release or
  vintage, follow-up cutoff, calendar grouping, and era-specific field mapping.
- In required_analysis_assumptions.statistical_model, record aggregation,
  smoothing, regression, joinpoint, survival, forecasting, significance, and
  model-selection settings needed for the target when the paper states them.
- Do not invent a protocol detail. If a necessary choice is not specified in
  the paper, add a concise statement in the relevant category beginning with
  "Not specified by the source:" so the ambiguity remains traceable.
- Neither required_analysis_assumptions nor agent_methodology_context may
  reveal the target answer, point estimate, comparison conclusion, or trend.
  Answer-bearing text belongs in evaluator_evidence.

Return only JSON with exactly one top-level key, "atomic_claims". Each item must contain exactly:
- "claim"
- "evidence_type"
- "answer": {"value", "unit", "label", "answer_type", "components", "uncertainty", "range_semantics"}
- "verification_target"
- "source_quote"
- "required_analysis_assumptions": {"cohort", "coding", "denominator_and_standardization", "temporal_data", "statistical_model"}
- "agent_methodology_context"
- "evaluator_evidence"

Each required numeric component must contain exactly "value", "unit", "label",
"role", "relation", "comparator", "required", "reported_decimal_places", and
"tolerance". tolerance contains exactly "kind", "reproduced", and "close".
Optional uncertainty components use the same fields but may set tolerance=null.
Relation must be one of "eq", "lt", "lte", "gt", or "gte".

For categorical or qualitative answers, set components=[], uncertainty=null,
and range_semantics=null.

Example estimate with confidence interval:
{"value":"2.9 (95% CI 2.7-3.1)","unit":"relative rate","label":"relative rate","answer_type":"continuous_numeric","components":[{"value":"2.9","unit":"relative rate","label":"point estimate","role":"primary","relation":"eq","comparator":"relative_numeric","required":true,"reported_decimal_places":1,"tolerance":{"kind":"relative","reproduced":0.05,"close":0.10}}],"uncertainty":{"kind":"confidence_interval","level":"95%","lower":{"value":"2.7","unit":"relative rate","label":"lower confidence limit","role":"range_lower","relation":"eq","comparator":"relative_numeric","required":false,"reported_decimal_places":1,"tolerance":null},"upper":{"value":"3.1","unit":"relative rate","label":"upper confidence limit","role":"range_upper","relation":"eq","comparator":"relative_numeric","required":false,"reported_decimal_places":1,"tolerance":null}},"range_semantics":null}

Example split for a before/after finding (two items in atomic_claims):
{"value":"4.9","unit":"per 100,000","label":"1975 baseline incidence","answer_type":"continuous_numeric","components":[{"value":"4.9","unit":"per 100,000","label":"1975 baseline incidence","role":"primary","relation":"eq","comparator":"relative_numeric","required":true,"reported_decimal_places":1,"tolerance":{"kind":"relative","reproduced":0.05,"close":0.10}}],"uncertainty":null,"range_semantics":null}
{"value":"14.3","unit":"per 100,000","label":"2009 follow-up incidence","answer_type":"continuous_numeric","components":[{"value":"14.3","unit":"per 100,000","label":"2009 follow-up incidence","role":"primary","relation":"eq","comparator":"relative_numeric","required":true,"reported_decimal_places":1,"tolerance":{"kind":"relative","reproduced":0.05,"close":0.10}}],"uncertainty":null,"range_semantics":null}

Example lower-bound finding, "the IRR reached at least 0.7":
{"value":"at least 0.7","unit":"IRR","label":"lower-bound IRR","answer_type":"continuous_numeric","components":[{"value":"0.7","unit":"IRR","label":"lower-bound IRR","role":"primary","relation":"gte","comparator":"relative_numeric","required":true,"reported_decimal_places":null,"tolerance":{"kind":"relative","reproduced":0.05,"close":0.10}}],"uncertainty":null,"range_semantics":null}

Example upper-bound finding, "the percentage did not exceed 9%":
{"value":"not exceeding 9","unit":"%","label":"upper-bound percentage","answer_type":"continuous_numeric","components":[{"value":"9","unit":"%","label":"upper-bound percentage","role":"primary","relation":"lte","comparator":"relative_numeric","required":true,"reported_decimal_places":null,"tolerance":{"kind":"relative","reproduced":0.05,"close":0.10}}],"uncertainty":null,"range_semantics":null}

Worked example:
Source finding: "Overall, the age-adjusted CRC incidence rates significantly
decreased between 1975 and 2010 (APC -0.92; 95% CI -1.14 to -0.70, P<0.001)."
Better atomic claim (with this paper's context): "Among histologically confirmed colon or rectal cancers
diagnosed in SEER 9 from 1975 through 2010, excluding nursing-home, autopsy, and
death-certificate reports, the age-adjusted incidence rate decreased by 0.92%
per year (95% CI -1.14% to -0.70%; two-sided P<0.001)."
Because this claim asks how much incidence decreased, use positive 0.92 as the
sole required answer component and label it "magnitude of annual incidence
decline". Store -1.14 and -0.70 as optional confidence bounds because those are
signed APC limits, and do not create a separate p-value claim. If the claim were
instead worded "the signed APC was -0.92%", preserve -0.92 as the target.

Paper metadata:
{{paper_metadata}}

Grouped source claim:
{{source_claim}}

Full paper text:
{{paper_context}}


Input fields:
  paper_metadata: {paper_id, pmid, pmcid, title, seer_registries, starting_question, source_insight_id}
  source_claim: one grouped finding, including its numeric inventory
  paper_context: full paper text

Output fields:
{
  atomic_claims: [{
    claim: string,
    evidence_type: string,
    answer: {
      value: string,
      unit: string,
      label: string,
      answer_type: "continuous_numeric" | "exact_numeric" | "count" | "categorical" | "qualitative",
      components: [{
        value: string,
        unit: string,
        label: string,
        role: "primary" | "comparison" | "supporting" | "range_lower" | "range_upper",
        relation: "eq" | "lt" | "lte" | "gt" | "gte",
        comparator: "relative_numeric" | "exact_numeric",
        required: boolean,
        reported_decimal_places: integer | null,
        tolerance: {
          kind: "relative" | "absolute" | "exact",
          reproduced: number,
          close: number
        } | null
      }],
      uncertainty: {
        kind: "confidence_interval" | "uncertainty_interval" | "interquartile_range" | "range",
        level: string,
        lower: {
          value: string,
          unit: string,
          label: string,
          role: "primary" | "comparison" | "supporting" | "range_lower" | "range_upper",
          relation: "eq" | "lt" | "lte" | "gt" | "gte",
          comparator: "relative_numeric" | "exact_numeric",
          required: boolean,
          reported_decimal_places: integer | null,
          tolerance: {
            kind: "relative" | "absolute" | "exact",
            reproduced: number,
            close: number
          } | null
        },
        upper: {
          value: string,
          unit: string,
          label: string,
          role: "primary" | "comparison" | "supporting" | "range_lower" | "range_upper",
          relation: "eq" | "lt" | "lte" | "gt" | "gte",
          comparator: "relative_numeric" | "exact_numeric",
          required: boolean,
          reported_decimal_places: integer | null,
          tolerance: {
            kind: "relative" | "absolute" | "exact",
            reproduced: number,
            close: number
          } | null
        }
      } | null,
      range_semantics: "reported_interval" | "acceptable_interval" | null
    },
    verification_target: string,
    source_quote: string,
    required_analysis_assumptions: {
      cohort: [string],
      coding: [string],
      denominator_and_standardization: [string],
      temporal_data: [string],
      statistical_model: [string]
    },
    agent_methodology_context: [string],
    evaluator_evidence: [string]
  }]
}
\end{lstlisting}

\begin{lstlisting}[style=seerprompt,caption={Reproduction execution},label={prompt:reproduction-execution}]
# Task: reproduce one published finding from the SEER 9 data

You are given direct query access to the SEER 9 cancer-registry table on a
SQL warehouse. Your job is to **independently reproduce the finding
below** by analyzing the data yourself (SQL to pull it, then any computation you
like), then report whether your computed result or observed pattern matches it.

## Temporal constraint (important)
The finding comes from a study published in **{{publication_year}}**. To make this a fair
reproduction of what that study could have done, restrict yourself to data and
resources that existed **on or before {{publication_year}}**:
- Do not use methods, software versions, or published results that postdate
  {{publication_year}}. The only data-release exceptions are the APPROVED staged population
  denominators and SEER*Explorer aggregate tables described below. Use those
  staged resources regardless of their packaging vintage, restrict their rows to
  the study's years and cohort, and do not go looking for an older vintage.
The SEER 9 case data itself is appropriate to use as provided.

## Do NOT look up the source (independence requirement)
This reproduction may use the provided source paper text for methodological
context, but the numbers must still be computed independently from the SEER 9
data. Therefore:
- You may read `paper_context.txt` to understand cohort definitions, exclusions,
  coding choices, statistical methods, tables/figure captions, and other context
  needed to reproduce the target claim.
- Do not search for, open, or fetch the paper, its abstract, or secondary pages
  from the web -- e.g. PubMed, PubMed Central (ncbi.nlm.nih.gov / pmc), journal
  sites, ResearchGate, or any page quoting the published numbers. Use the staged
  `paper_context.txt` copy only.
- Use the STAGED population tables for denominators (see "Population denominators"
  below) -- do NOT fetch population data from the web. The only acceptable
  web/reference use is generic, non-paper methodology documentation (e.g. how
  age-adjustment / APC works).
- The target claim below is given ONLY so you can state whether you matched or
  verified it. Treat any target numbers as a checksum -- do NOT copy results out
  of `paper_context.txt` or reverse-engineer your cohort/method to land on them.
  Use the paper for methods/context, compute or inspect the result yourself, then
  compare.

## The finding to reproduce
{{finding_json}}

## Paper context
The full source paper text is available in `paper_context.txt`. Use it as
context for the study's cohort, variable definitions, exclusions, and statistical
methods. The final reproduced number must come from your own data analysis.

## Local SEER manuals and documentation
Official NCI/SEER documentation is available under `manuals/`.
Start with `manuals/README.md`, which maps each local file to its source URL and
summarizes the contents. Useful paths include:
- `manuals/seerstat_nov2025/`: SEER*Stat variable dictionary and Research /
  Research Plus data description.
- `manuals/population/`: population denominator dictionary, modifications,
  single-year-age methods, hurricane-adjustment notes, and standard-population
  references.
- `manuals/coding_manuals/`: SEER Program Coding and Staging Manuals plus
  county, state, race, and reportability appendices.
- `manuals/variable_recodes/`: site recodes, stage variables, cause-of-death
  recodes, cause-specific death classification, race/ethnicity, months survived,
  and tumor-size recodes.
- `manuals/registry/`: SEER registry lists, groupings, and population
  characteristics.

Prefer these local manuals over web search for SEER documentation. You can use
`rg` and `find` to locate HTML/README content; for PDFs, use `pdftotext`, e.g.
`pdftotext manuals/seerstat_nov2025/TextData.FileDescription-nov2025.pdf - | rg -i "summary stage"`.
For the XLSX variable dictionary, use Python/pandas/openpyxl if the PDF or HTML
copy is not enough.

## Official SEER*Explorer aggregate tables
The warehouse also contains 34 staged tables from the official November 2024
SEER*Explorer archive. Their fully qualified names and columns are listed in
`schema.txt`; all names begin with `{{catalog}}.{{schema}}.seer_explorer_`.
These tables provide published aggregate incidence, mortality, survival,
prevalence, lifetime-risk, rural/urban, and preliminary-estimate data.

Read `manuals/seer_explorer_footnotes.md` before using an Explorer table. It
records data sources, age-adjustment and trend methodology, coding caveats, and
tab-specific qualifications. These are approved staged data resources even
though the archive packaging is a current release. Restrict queries to the
study's years, population, site, sex, race/ethnicity, age, and stage definitions,
and state clearly when an Explorer aggregate rather than the case-level table
supports the result.

## How to get the data
Read the SEER 9 table with the provided tool:

    python seer_sql.py "SELECT ... FROM {{case_table}} WHERE ..."

- See `schema.txt` for all column names and notes on the all-STRING typing and
  missing-value sentinels.
- The table is `{{case_table}}`.

## Population denominators -- use ONLY the staged tables
The SEER-9 population denominator tables in the database (described in
`schema.txt`, `{{catalog}}.{{schema}}.seer_pop_*`) are the APPROVED, authoritative
denominators for this task. They cover **1969-2024** with one row per
year x county x age x sex x race, so restrict to the years you need with a
`WHERE year BETWEEN ...` clause.
- Do NOT fetch external population files (SEER popdata, Census, NBER, etc.) over
  the web or in your own code -- use the staged tables as-is.
- These tables are a current-release vintage. The small population-estimate
  revision vs. the vintage the original study used is well within tolerance and
  is NOT worth pursuing -- do not hunt for an older-vintage file.
- Restrict the population to the SEER-9 counties (per the FIPS map in
  `schema.txt`) so the denominator matches the case cohort.

## How to compute -- work however you like
You are NOT limited to SQL. Pull whatever data you need with `seer_sql.py`, then
analyze it however is most appropriate:
- numpy, scipy, pandas, and statsmodels are PRE-INSTALLED -- just `import` them
  (no pip needed, and network installs will fail). Use them to save intermediate
  CSVs and fit regressions (age-adjustment, trend/APC models).
- There is no NCI Joinpoint binary here; implement joinpoint-style APC as a
  segmented / piecewise log-linear regression with statsmodels (with its standard
  errors / CIs) rather than a hand-rolled fit.
- Create as many working files in this directory as you want; you have a full
  scratch workspace here.
- You can also choose to just stick with SQL.

## Ground rules
- **Read-only on the database only.** The SEER table is shared and must not be
  modified: no INSERT/UPDATE/DELETE, DROP, CREATE/ALTER TABLE, or new tables/views
  -- issue **SELECT queries only** against it. This is the one hard restriction.
- Everything else (local files, scripts, packages) is fair game.

## What to deliver
Write your final report to a file named exactly `report.md` in this directory.
It must contain:
1. How you computed it: the SQL you pulled, plus any Python/analysis you ran. Just put all code under code blocks inside the markdown file.
2. The numbers, pattern, figure/table property, or context item you computed,
   inspected, or applied.
3. A clear verdict: does your computation reproduce the finding (and how closely)?
4. If it does not match, your best explanation why (cohort definition, coding,
   age-adjustment standard, SEER 9 vs other registry coverage, etc.).

Only `report.md` is collected after the run; scratch files you create elsewhere
are discarded.

Input fields:
  publication_year: source publication year
  finding_json: one grouped finding
  catalog, schema, table: database identifiers
  Files: insight.json, paper_context.txt, schema.txt, manuals/, seer_sql.py

Output fields:
  report.md: string (Markdown report)
    Contents: computation code; computed or inspected results; reproduction verdict; explanation of discrepancies.
\end{lstlisting}

\begin{lstlisting}[style=seerprompt,caption={Atomic reproduction grading},label={prompt:atomic-grading}]
You are grading whether a data-analysis agent
reproduced EACH atomic result derived from one grouped SEER finding.

The report was originally produced for the grouped finding. Inspect every
atomic target independently. A value or conclusion for one target must not
affect any other target.

Your job is evidence extraction and method assessment, not numeric grading:
- answer_status is "reported" when the report gives the target result,
  "not_reported" when it does not, and "unclear" when the evidence is ambiguous.
- method_status is "valid" when the reported result was produced with a sound
  method for this target, "invalid" for a materially wrong cohort, measure,
  denominator, grouping, or calculation, and "unclear" when the report lacks
  enough methodological detail.
- Treat required_analysis_assumptions as part of the target protocol when
  assessing method_status. The agent report does not need to repeat every
  assumption verbatim, but its computation must be compatible with applicable
  cohort bounds, coding rules, denominator and standardization conventions,
  temporal data definitions, and statistical-model settings. A conflicting
  assumption is method_status "invalid"; insufficient evidence is "unclear".
- if_cheating is true only when the report indicates that the agent obtained
  the target result by recalling it, copying it from the source paper, or
  searching online for the reported answer instead of computing it from the
  provided data. Looking up documentation, variable definitions, or coding
  guidance is allowed when the agent then performs the computation itself.
  Do not infer cheating merely because a computed value matches the target.
  When the report does not establish a credible computation, use
  method_status "unclear" or "invalid" as appropriate even if
  if_cheating is false.
- For numeric targets, return one object per corresponding required result in
  computed_components. Each object has value, relation, and unit. Use the
  report's unrounded computed value whenever it is available, in target order
  and target units. Most targets require one number. A reported interval requires
  two endpoints; an acceptable interval requires the one computed scalar being
  tested. Never duplicate one scalar to manufacture two endpoints. Also put a
  compact legacy rendering of these same values in computed_value. Ignore
  optional confidence bounds. Set semantic_verdict to "not_applicable". Do not
  calculate relative error, apply a tolerance, or assign a final verdict.
- Normalize change direction to the representation requested by the atomic
  target. If the target asks for the magnitude of a decline or says "decreased
  by", convert a report's signed APC of -1.09 to the positive magnitude 1.09.
  If the target asks for a signed APC, preserve -1.09. Do not otherwise change
  signs.
- For a lower- or upper-bound target, extract the report's actual computed value,
  not the target boundary. Example: for target ">5", return "5.51" when the
  report computed 5.51. For target "P<0.0001", return "2.94e-26" when that is
  the computed P value. If the report supplies only a bound, preserve its
  relation in computed_components and operator in computed_value, such as
  "<0.0001". Never replace a reported value with the reference threshold.
- Preserve the unit and scale actually computed. For example, distinguish a
  percentage from a proportion and a rate per 100,000 from a raw count. Use an
  empty unit only when the report truly does not state or establish one.
- For qualitative or categorical targets, set semantic_verdict to
  "equivalent", "approximately_equivalent", "conflicting", or "unclear" by
  comparing the report's conclusion with the target. Use "not_applicable" only
  for numeric targets.

Return ONLY one JSON object with key "grades". "grades" must contain exactly
one object for every supplied atomic_claim_id, with no missing or extra IDs.
Each grade object must contain exactly these keys: atomic_claim_id,
answer_status, computed_components, computed_value, method_status, if_cheating,
semantic_verdict, key_caveat, and rationale. computed_components must be [] for
non-numeric targets. Each numeric computed component must have exactly value,
relation, and unit, where relation is eq, lt, lte, gt, or gte. if_cheating must
be a JSON boolean. Use only the enum values specified above.

## ATOMIC targets and grading policies
{{targets}}

## Agent REPORT
{{report}}


Input fields:
  targets: [{atomic_claim_id, claim, answer, evaluation_mode, verification_target, source_quote, required_analysis_assumptions}]
  report: reproduction report

Output fields:
{
  grades: [{
    atomic_claim_id: string,
    answer_status: "reported" | "not_reported" | "unclear",
    computed_components: [{
      value: string,
      relation: "eq" | "lt" | "lte" | "gt" | "gte",
      unit: string
    }],
    computed_value: string,
    method_status: "valid" | "invalid" | "unclear",
    if_cheating: boolean,
    semantic_verdict: "equivalent" | "approximately_equivalent" | "conflicting" | "unclear" | "not_applicable",
    key_caveat: string,
    rationale: string
  }]
}
\end{lstlisting}

\begin{lstlisting}[style=seerprompt,caption={Reproduction failure classification},label={prompt:reproduction-failure-classification}]
You are analyzing failed atomic targets from a scientific reproduction agent that queried staged NCI SEER data. Classify why each target was not reproduced.

Use the reproduction report and grader records as evidence. The grader record contains the target, extracted computed value, and prior rationale, but you must distinguish the underlying cause rather than restating the numeric mismatch.

Assign exactly one primary_issue per atomic target:
- data_unavailable: required records, variables, geography, endpoint, years, or historical schema are absent from the staged data.
- target_under_specified: the paper's exact operational definition, model, test, or context cannot be uniquely inferred from the target and available documentation.
- agent_omission: the target appears computable, but the agent did not attempt it or failed to report the result.
- cohort_or_coding_error: the agent used the wrong population filter, disease/histology definition, stage, treatment, or tumor-size coding.
- denominator_or_standardization_error: the agent used the wrong denominator, rate scale, age standard, weighting, or change formula.
- statistical_method_error: the agent used an unsuitable test, regression, joinpoint/time segmentation, aggregation, or other statistical procedure.
- survival_or_endpoint_error: the agent used the wrong mortality/survival endpoint, censoring, event definition, or follow-up.
- valid_alternative_estimate: the agent performed a defensible computation, but a data-vintage, extract, small-sample, or reasonable operationalization difference produced a materially different estimate.
- qualitative_result_conflict: the computed direction, ranking, or pattern directly conflicts with the target and no narrower cause is supported.
- grading_artifact: the substantive result should likely count, but atomicization, parsing, rounding, inequality, or tolerance policy caused rejection.
- contamination_or_lookup: the answer was copied, recalled, or obtained online instead of being computed from staged data.
- insufficient_evidence: evidence is inadequate to choose another category.

issue_codes are narrower contributing diagnoses. Put the code that best supports primary_issue first. Multiple codes are allowed only when independently supported.

Use only these exact issue_codes strings:
missing_records, missing_variable, missing_geographic_granularity, missing_external_endpoint, temporal_coverage_or_schema, paper_method_not_identifiable, claim_context_incomplete, not_attempted, result_not_reported, wrong_cohort_filter, wrong_disease_or_histology_coding, wrong_stage_treatment_or_size_coding, wrong_denominator, wrong_age_standardization, wrong_rate_or_change_calculation, wrong_statistical_test, wrong_model_or_time_segmentation, small_sample_instability, wrong_survival_endpoint, wrong_censoring_or_followup, data_vintage_or_extract_difference, defensible_operationalization_difference, qualitative_direction_or_ranking_conflict, numeric_tolerance_or_rounding, atomic_target_or_reference_issue, copied_or_recalled_answer, tool_or_query_failure, cause_unclear.

Use only these exact responsibility strings: benchmark_data, benchmark_specification, agent, evaluator, scientific_variation, unclear.

Use only these exact computability strings: computable_from_staged_data, partially_computable, not_computable_from_staged_data, unclear.

Use only the strings high, medium, or low for confidence. The evidence field must be one string, not a list.

Important distinctions:
- Missing data is not an agent error when the report correctly analyzes and documents the absence.
- A value outside tolerance is not automatically an agent error. Use valid_alternative_estimate when the method is defensible and comparable.
- Use grading_artifact for values that agree after the benchmark's intended rounding/inequality semantics or for a defective atomic reference.
- Qualitative grading is intentionally binary. A qualitative grade of close_to_reproduced or approximately_equivalent is therefore an effective failure, not automatically a grading artifact. Diagnose the missing, weakened, or conflicting semantic content. Use grading_artifact only if the report actually establishes the complete target and the judge or reference rejected it incorrectly.
- Use target_under_specified only when a hidden paper-specific choice prevents unique reproduction, not merely because the agent chose poorly.
- Use agent_omission when data and method appear available but no result was attempted or reported.
- Documentation or web lookup is allowed when it supports a real computation. contamination_or_lookup requires that the reported result itself was not computed.
- Keep diagnosis and evidence specific to this case. Do not invent unavailable facts.

Return JSON with one classifications item for every input atomic_claim_id and no others. Do not create any enum values beyond those listed above. Each item must contain:
atomic_claim_id, primary_issue, issue_codes, responsibility, computability, diagnosis, evidence, recommended_action, confidence.

FAILED TARGETS:
{{failure_rows_json}}

REPRODUCTION REPORT:
{{report}}

Input fields:
  failure_rows_json: [{atomic_claim_id, claim, target_value, computed_value, rationale, grading_method, judge_verdict, numeric_verdict, if_cheating}]
  report: reproduction report

Output fields:
{
  classifications: [{
    atomic_claim_id: string,
    primary_issue: "data_unavailable" | "target_under_specified" | "agent_omission" | "cohort_or_coding_error" | "denominator_or_standardization_error" | "statistical_method_error" | "survival_or_endpoint_error" | "valid_alternative_estimate" | "qualitative_result_conflict" | "grading_artifact" | "contamination_or_lookup" | "insufficient_evidence",
    issue_codes: ["missing_records" | "missing_variable" | "missing_geographic_granularity" | "missing_external_endpoint" | "temporal_coverage_or_schema" | "paper_method_not_identifiable" | "claim_context_incomplete" | "not_attempted" | "result_not_reported" | "wrong_cohort_filter" | "wrong_disease_or_histology_coding" | "wrong_stage_treatment_or_size_coding" | "wrong_denominator" | "wrong_age_standardization" | "wrong_rate_or_change_calculation" | "wrong_statistical_test" | "wrong_model_or_time_segmentation" | "small_sample_instability" | "wrong_survival_endpoint" | "wrong_censoring_or_followup" | "data_vintage_or_extract_difference" | "defensible_operationalization_difference" | "qualitative_direction_or_ranking_conflict" | "numeric_tolerance_or_rounding" | "atomic_target_or_reference_issue" | "copied_or_recalled_answer" | "tool_or_query_failure" | "cause_unclear"],
    responsibility: "benchmark_data" | "benchmark_specification" | "agent" | "evaluator" | "scientific_variation" | "unclear",
    computability: "computable_from_staged_data" | "partially_computable" | "not_computable_from_staged_data" | "unclear",
    diagnosis: string,
    evidence: string,
    recommended_action: string,
    confidence: "high" | "medium" | "low"
  }]
}
\end{lstlisting}

\lstinputlisting[style=seerprompt,caption={Scientific frame and eligible claims},label={prompt:paper-frame}]{appendix_assets/prompts/07_paper_frame.txt}

\begin{lstlisting}[style=seerprompt,caption={Candidate choices in generated reports},label={prompt:artifact-candidates}]
Extract scientifically material analytical assumptions from one generated report and its own original query log. Use the supplied canonical paper claims only for linkage; do not force the report to address every claim. Do not use or predict sensitivity outcomes.

Candidate IDs must use candidate_prefix followed by sequential two-digit numbers. One candidate must represent one underlying decision. Distinguish explicit report statements, execution-observed settings, necessary but unspecified choices, and genuinely unknown decisions. Do not treat absent analysis as an assumption. Evidence quotes must be exact contiguous substrings of report or query_log. Return only the JSON object required by the schema.

Input fields:
  artifact: source artifact identity and metadata
  canonical_claims: claims from paper_frame
  report: string
  query_log: string
  observability_limitations: string
  assumption_families: array of the ten family names
  candidate_prefix: "S" for the Sol report; "O" for the Opus report
  outcome_blindness: string

Output fields:
{
  artifact_id: string,
  source: "codex" | "opus",
  candidates: [{
    candidate_id: string,
    family: "population" | "measurement" | "time" | "missingness" | "denominator" | "endpoint" | "model" | "inference" | "subgroup" | "interpretation",
    name: string,
    decision: string,
    original_value: string,
    observability: "explicit" | "execution_observed" | "implicit_unspecified" | "unknown",
    linked_claim_ids: [string],
    applicability: string,
    consequence_mechanism: string,
    evidence: [{source: "report" | "query_log" | "reproduction_index", quote: string, location: string}]
  }],
  coverage_gaps: [string]
}
\end{lstlisting}

\lstinputlisting[style=seerprompt,caption={Candidate extraction repair},label={prompt:artifact-candidates-repair}]{appendix_assets/prompts/09_artifact_candidates_repair.txt}

\lstinputlisting[style=seerprompt,caption={Shared analytical choice selection},label={prompt:canonical-panel}]{appendix_assets/prompts/10_canonical_panel.txt}

\lstinputlisting[style=seerprompt,caption={Analytical choice review},label={prompt:panel-review}]{appendix_assets/prompts/11_panel_review.txt}

\lstinputlisting[style=seerprompt,caption={Analytical choice repair},label={prompt:panel-repair}]{appendix_assets/prompts/12_panel_repair.txt}

\lstinputlisting[style=seerprompt,caption={Final analytical choice repair},label={prompt:panel-repair-final}]{appendix_assets/prompts/13_panel_repair_final.txt}

\lstinputlisting[style=seerprompt,caption={Final analytical choice review},label={prompt:panel-review-final}]{appendix_assets/prompts/14_panel_review_final.txt}

\lstinputlisting[style=seerprompt,caption={Analytical choice resolution},label={prompt:panel-resolution}]{appendix_assets/prompts/15_panel_resolution.txt}

\lstinputlisting[style=seerprompt,caption={Source-specific baseline mapping},label={prompt:source-mapping}]{appendix_assets/prompts/16_source_mapping.txt}

\lstinputlisting[style=seerprompt,caption={Source mapping repair},label={prompt:source-mapping-repair}]{appendix_assets/prompts/17_source_mapping_repair.txt}

\lstinputlisting[style=seerprompt,caption={Shared alternative generation},label={prompt:shared-alternatives}]{appendix_assets/prompts/18_shared_alternatives.txt}

\lstinputlisting[style=seerprompt,caption={Alternative generation repair},label={prompt:shared-alternatives-repair}]{appendix_assets/prompts/19_shared_alternatives_repair.txt}

\lstinputlisting[style=seerprompt,caption={Prospective alternative review},label={prompt:alternative-review}]{appendix_assets/prompts/20_alternative_review.txt}

\lstinputlisting[style=seerprompt,caption={Alternative review repair},label={prompt:alternative-review-repair}]{appendix_assets/prompts/21_alternative_review_repair.txt}

\lstinputlisting[style=seerprompt,caption={Execution planning},label={prompt:execution-plan}]{appendix_assets/prompts/22_execution_plan.txt}

\lstinputlisting[style=seerprompt,caption={Execution plan repair},label={prompt:execution-plan-repair}]{appendix_assets/prompts/23_execution_plan_repair.txt}

\lstinputlisting[style=seerprompt,caption={Analysis execution},label={prompt:analysis-execution}]{appendix_assets/prompts/24_audit_execution.txt}

\lstinputlisting[style=seerprompt,caption={Consequence adjudication},label={prompt:adjudication}]{appendix_assets/prompts/25_adjudication.txt}

\lstinputlisting[style=seerprompt,caption={Execution continuation},label={prompt:executor-continuation}]{appendix_assets/prompts/26_executor_continuation.txt}

\lstinputlisting[style=seerprompt,caption={Execution finalization},label={prompt:executor-finalization}]{appendix_assets/prompts/27_executor_finalization.txt}

\lstinputlisting[style=seerprompt,caption={Execution task},label={prompt:executor-launch}]{appendix_assets/prompts/28_executor_launch.txt}

\begin{lstlisting}[style=seerprompt,caption={Structured output repair},label={prompt:structured-output-repair}]
Return a corrected complete JSON object matching the supplied schema. Preserve supported scientific content, fix the validation error, and do not add commentary. Validation error: {{validation_error}}

Input fields:
  schema: JSON Schema defining output fields, nested structure, and types
  inputs: task input values
  Previous assistant response
  validation_error: parsing or schema error

Output fields:
  object (field names, nested structure, and value types defined by the supplied schema)
\end{lstlisting}

\lstinputlisting[style=seerprompt,caption={One-shot analytical-choice analysis},label={prompt:one-shot-consequential-analysis}]{appendix_assets/prompts/30_one_shot_consequential_audit.txt}

\lstinputlisting[style=seerprompt,caption={Ordinal consequentiality grading},label={prompt:consequentiality-ordinal-v3}]{appendix_assets/prompts/31_consequentiality_ordinal_v3.txt}

\lstinputlisting[style=seerprompt,caption={Mapping high-score analyses to individual findings and conclusions of multiple findings},label={prompt:finding-consequence-mapping}]{appendix_assets/prompts/32_finding_consequence_mapping.txt}

\end{document}